\documentclass[10pt,twocolumn,letterpaper]{article}
\usepackage[pagenumbers]{cvpr} 

\usepackage{multirow}
\usepackage{booktabs} 
\usepackage{amsmath,amssymb}  
\usepackage{graphicx}
\usepackage{cuted}
\usepackage{capt-of}
\usepackage{algorithm}
\usepackage{algpseudocode}
\usepackage{etoc}
\usepackage{gensymb}
\usepackage{makecell}
\usepackage{dsfont}
\usepackage[accsupp]{axessibility}

\renewcommand{\paragraph}[1]{\vspace{.5em}\noindent\textbf{#1.}}

\usepackage{xspace}
\newcommand{\ours}{InterPrior\xspace}

\definecolor{cvprblue}{rgb}{0.21,0.49,0.74}
\usepackage[pagebackref,breaklinks,colorlinks,allcolors=cvprblue]{hyperref}

\title{\ours: Scaling Generative Control for Physics-Based \\ Human-Object Interactions}

\author{Sirui Xu$^{1}$ \quad Samuel Schulter$^{2}$ \quad Morteza Ziyadi$^{2}$ \quad Xialin He$^{1}$ \\ Xiaohan Fei$^{2}$ \quad
Yu-Xiong Wang$^{1\dag}$ \quad
Liang-Yan Gui$^{1\dag}$\\
$^{1}$ University of Illinois Urbana-Champaign \quad $^{2}$ Amazon\\
$^{\dag}$ Equal Advising\\
\small\url{https://sirui-xu.github.io/InterPrior}}
\begin{document}
\maketitle
\begin{strip}\centering
\vspace{-4.5em}
\includegraphics[width=\textwidth]{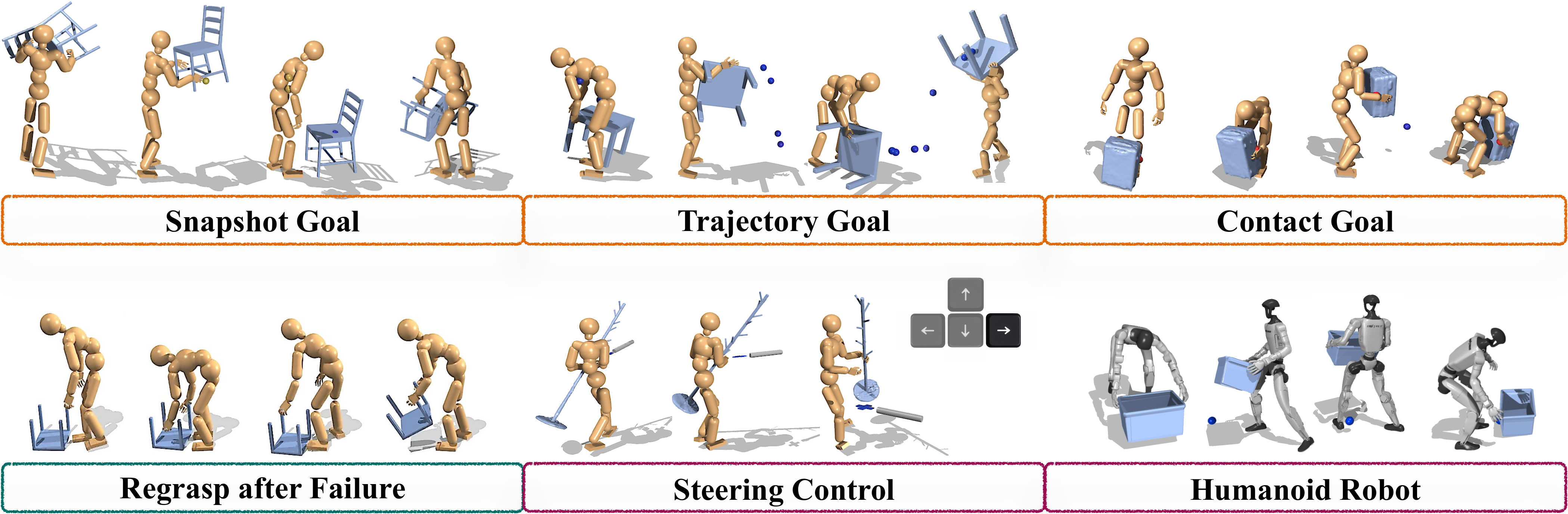}
\captionof{figure}{\ours is a \textit{versatile generative controller} instantiated as a goal-conditioned policy that controls a simulated humanoid to follow goal guidance and interact with objects in a physics-based simulator. Three core, composable capabilities enable pursuing (\textbf{I}) long-horizon snapshot goals, (\textbf{II}) trajectory goals, and (\textbf{III}) contact goals (\textit{Top}). \textcolor[rgb]{0.85,0.7,0.0}{\textit{Yellow}}, \textcolor[rgb]{0.0,0.2,0.6}{\textit{blue}}, and \textcolor[rgb]{0.7,0.0,0.0}{\textit{red}} dots respectively denote \textcolor[rgb]{0.85,0.7,0.0}{\textit{human}}, \textcolor[rgb]{0.0,0.2,0.6}{\textit{object}}, and \textcolor[rgb]{0.7,0.0,0.0}{\textit{contact}} goals. It demonstrates failure recovery (\textit{Bottom Left}) from unsuccessful grasps. \ours enables steering control from a human operator and can be applied to humanoid robot embodiments (\textit{Bottom Right}). More demo videos are provided in the \href{https://sirui-xu.github.io/InterPrior}{webpage}.
\label{fig:teaser}}
\end{strip}
\etocdepthtag.toc{mtchapter}
\begin{abstract}
Humans rarely plan whole-body interactions with objects at the level of explicit whole-body movements. High-level intentions, such as affordance, define the goal, while coordinated balance, contact, and manipulation can emerge naturally from underlying physical and motor priors. Scaling such priors is key to enabling humanoids to compose and generalize loco-manipulation skills across diverse contexts while maintaining physically coherent whole-body coordination. To this end, we introduce InterPrior, a scalable framework that learns a unified generative controller through large-scale imitation pretraining and post-training by reinforcement learning. InterPrior first distills a full-reference imitation expert into a versatile, goal-conditioned variational policy that reconstructs motion from multimodal observations and high-level intent. While the distilled policy reconstructs training behaviors, it does not generalize reliably due to the vast configuration space of large-scale human-object interactions. To address this, we apply data augmentation with physical perturbations, and then perform reinforcement learning finetuning to improve competence on unseen goals and initializations. Together, these steps consolidate the reconstructed latent skills into a valid manifold, yielding a motion prior that generalizes beyond the training data, e.g., it can incorporate new behaviors such as interactions with unseen objects. We further demonstrate its effectiveness for user-interactive control and its potential for real robot deployment.
\end{abstract}    
\section{Introduction}
\label{sec:intro}
Human-object interaction (HOI) is inherently hierarchical: humans plan at a high level with sparse intentions, while detailed limb coordination, balance, and contact emerge through \textit{fast}, \textit{intuitive} motor responses~\cite{todorov2002optimal}. For instance, when reaching for a bottle, we plan the hand’s target and object motion, while the rest of the body follows through \textit{subconscious} coordination.
Motion imitation policies~\cite{xu2025intermimic} have scaled to large HOI skills but rely on explicit planners for dense full-body and object references. In contrast, an \textit{interaction motor prior} should sample feasible loco-manipulation behaviors from a distribution conditioned on sparse goals, \eg, next-second hand contact, rather than simply mimicking deterministic, fully specified trajectories.

To model a distribution over feasible loco-manipulation behaviors, early work~\cite{hassan2023synthesizing,pan2025tokenhsi} learns a generative controller via adversarial distributional matching and then uses reinforcement learning (RL) to promote task achievement under it. These methods can expand motion coverage beyond demonstrations, but are hard to scale due to unstable optimization, discriminator mode collapse, and handcrafted task objectives. An alternative is to distill reference imitation policies~\cite{luo2024grasping}, with goal conditioning~\cite{tessler2025maskedmanipulator} achieved without task-specific design. While these approaches can absorb large-scale data, they can be brittle when reference coverage lags far behind the configuration space—as in loco-manipulation, where even a few object degrees of freedom can induce a combinatorial explosion of contact modes and relative poses with different geometries.

To address these limitations, we introduce \emph{\ours}, a physics-based HOI controller that is \emph{scalable} along four axes (Figure~\ref{fig:teaser}). (\textbf{I}) \textit{task coverage}: a single policy supports multiple goal formulations, \eg, sparse targets and their compositions; (\textbf{II}) \textit{skill coverage}: the same training recipe scales to large HOI data and enables affordance-rich interactions beyond simple grasping; (\textbf{III}) \textit{motion coverage}: it generates expressive trajectories instead of merely reconstructing demonstrations; and (\textbf{IV}) \textit{dynamics coverage}: it maintains task success under varied physical properties.

Our key insight is that \textit{RL finetuning} is essential for turning distillation from data reconstruction into a robust, generalizable policy. Distillation alone cannot cover the full HOI configuration space, yet RL applied in isolation often drifts toward unnatural reward-hacking behaviors. We therefore use distillation to provide a strong, natural initialization, and apply RL as a \emph{local optimizer} that improves robustness while remaining anchored to the pretrained model.
Concretely, we leverage \textit{distillation} to inherit broad skills from large-scale HOI demonstrations, by training a masked conditional variational policy to reconstruct motor control from sparse, multimodal goals, distilled from a reference imitation expert. We then \textit{RL finetune} this policy to consolidate its latent skills into a \emph{valid interaction manifold}. The finetuning optimizes two objectives: improving success on unseen goals and initializations, and preserving pretrained knowledge through regularization. It leverages the pretrained base policy to synthesize natural in-between motions, with failure states to acquire recovery behaviors, \eg re-approach and re-grasp. Together, these steps transform reconstructed latent skills into a stable, continuous manifold that generalizes beyond the training trajectories.

Our contributions are fourfold. (\textbf{I}) We present \emph{InterPrior}, a generalizable generative controller for physics-based human-object interaction, encompassing diverse skills rather than fixed procedural routines (\eg, approach, grasp, place) typical of prior work. (\textbf{II}) We develop an RL finetuning strategy that enables robust failure recovery and goal execution across varied configurations while maintaining human-like coordination. The resulting controller supports mid-trajectory command switching, re-grasps after failures, and remains stable under perturbations. 
(\textbf{III}) We show that our finetuning strategy naturally extends to 
\textit{novel objects and interactions}, functioning as a reusable prior.
(\textbf{IV}) We demonstrate embodiment flexibility by training on the G1 humanoid~\cite{unitreeg1} with sim-to-sim evaluation and enabling real-time control via keyboard interfaces.

\section{Related Work}
\label{sec:related}
Data-driven human interaction animation has progressed from kinematic models assuming simplified object dynamics~\cite{zhang2024scenic, wang2021scene, Zhao:ICCV:2023} to methods generating whole-body motions with dynamic objects~\cite{ghosh2022imos, wu2024human, jiang2024scaling, jiang2024autonomous, lu2024choice, xu2023interdiff, peng2023hoi, diller2023cg, david, he2024syncdiff, wu2025hoi,zhang2025interactanything, xue2025guiding,cong2025semgeomo,zeng2025chainhoi,petrov2025tridi,jia2025primhoi, geng2025auto, petrov2025echo, xu2024interdreamer}.
However, these kinematic approaches often exhibit implausible contact drift and interpenetration. Such limitations partly arise from existing HOI datasets~\cite{bhatnagar22behave, jiang2022chairs, huang2022intercap, zhang2023neuraldome, li2023object, zhao2023im, kim2024parahome, wu2024himo, zhang2024core4d, xie2024intertrack, zhang2024hoi, zhang2024force, lu2025humoto,xu2025perceiving}, which contain spatial or physical inconsistencies that impede the learning of realistic interactions.
Physics-based methods seek to address this gap but often rely on early curated datasets~\cite{taheri2020grab} focusing on limited yet high-fidelity hand-centric manipulations~\cite{tessler2025maskedmanipulator, wang2023physhoi, luo2024grasping}.
Recent advances in humanoid hardware~\cite{sun2025ulc, baek2025whole, zhao2025resmimic, kalaria2025dreamcontrol, fu2025demohlm} have begun to bridge the virtual and physical domains, though typically without too much agility.
Together, these developments highlight the need for \emph{scalable HOI priors}, models capable of generalizing across tasks, remaining robust to imperfect data, and synthesizing physically realistic HOIs.

\subsection{Physics-based Character Animation}
Physics-based character animation learns simulated controllers via RL, \eg, tracking reference motions~\cite{peng2018deepmimic, zhang2025add}. Scalability has been improved through multi-clip trackers with reference planners~\cite{wang2020unicon, won2020scalable, juravsky2024superpadl} without or with closed-loop schemes~\cite{tevet2024closd, xu2025parc}.
Nevertheless, such controllers remain constrained by their reference motion planners, making them fragile when the planned motions are dynamically unstable, a very common issue in HOI, where kinematic planners often neglect physical feasibility.
\emph{Learned generative priors} address this limitation by encoding physically plausible motor memory encoded into policies. One line of research employs adversarial imitation with discriminators~\cite{peng2021amp} to learn the motor prior, and later extends to skill embeddings~\cite{peng2022ase} and conditional control~\cite{tessler2023calm, dou2023c}. These approaches promote motion diversity but remain sample-inefficient and challenging to scale.
A complementary line distills motor skills into compact latent codes. Earlier work adopts model learning to train a variational autoencoder (VAE)~\cite{kingma2013auto} based controller~\cite{won2022physics, yao2022controlvae, yao2023moconvq, fussell2021supertrack}, while recent studies pretrain universal trackers~\cite{luo2023perpetual} and distill them into latent priors~\cite{luo2023universal}, masked policies~\cite{tessler2024maskedmimic}, or offline training with diffusion models~\cite{truong2024pdp, huang2025diffuse,wu2025uniphys}. Yet, these methods are often limited by the expert converage.
Our \ours synergizes the strength of both lines: it first distills large-scale motion imitators and finetunes it via RL, bridging a generative controller with versatile conditions while enhancing the control by alleviating out-of-distribution brittleness.

\subsection{Physics-based Human-Object Interaction}
Advances in physics-based character control have progressively expandeded the scope of HOI animation. Early approaches primarily focus on simple object dynamics, such as striking or sitting~\cite{peng2022ase, cui2024anyskill, chao2021learning, pan2023synthesizing, xiao2024unified}, whereas recent developments have extended to complex, scenario-specific sports and games~\cite{luo2024smplolympics, liu2017learning, xie2022learning, zhang2023learning, wang2023physhoi, wang2024strategy, bae2023pmp, wang2025hil}. Progress has also been observed in generalizable tasks, such as object carrying and rearrangement~\cite{zhang2023simulation, pan2025tokenhsi, wang2024sims, gao2024coohoi, hassan2023synthesizing, deng2025human, zhang2025humanoidverse, merel2020catch, shen2025detach, li2025learning}, predominantly enabled by adversarial imitation learning, while most systems remain skill-specific, relying on fixed procedural routines (\eg approach, grasp, place with regular-shaped objects). They struggle to adapt to objects that require careful affordances and fine-grained interaction skills (\eg, grasping a chair bar with one hand). To address these limitations, HOI motion imitation~\cite{xie2023hierarchical, wu2024human, yu2025skillmimic, xu2025intermimic} has emerged as a promising paradigm for scaling skill repertoires and capturing fine-grained interactions, as it directly emphasizes precision and stability. Distilling such imitation policies therefore represents a crucial step toward establishing a \emph{versatile HOI controller}.
However, existing efforts often exhibit narrow task coverage, emphasizing single-object proficiency~\cite{yu2025skillmimic} or relying on curated dataset with low-dynamic and hand-centric skills~\cite{tessler2025maskedmanipulator, luo2024grasping, luo2025emergent}. Our \ours provides a principled solution for generalizing a generative controller for agile whole-body loco-manipulation.
\section{Methodology}
\noindent\textbf{Task Formulation.}
We aim to learn a policy \(\pi\) that operates in a physics simulator and produces human-object interaction motion from high-level goals rather than full reference. Such goals can be extracted from a human user (\eg, steering control), a HOI kinematic motion generator (see Sec.~\ref{sec:add_exp}), or keypoints from Motion Captured (MoCap) data. The policy \(\pi\) conditions on the current human-object state and recent history together with these goals, and samples control signals from its learned distribution to drive the simulated human or humanoid to interact with the object. The outcome is a rollout motion sequence that is physically simulated, follows the provided goals where available, and remains diverse and natural in aspects that are not specified. 

\begin{figure}
    \centering
    \includegraphics[width=\columnwidth]{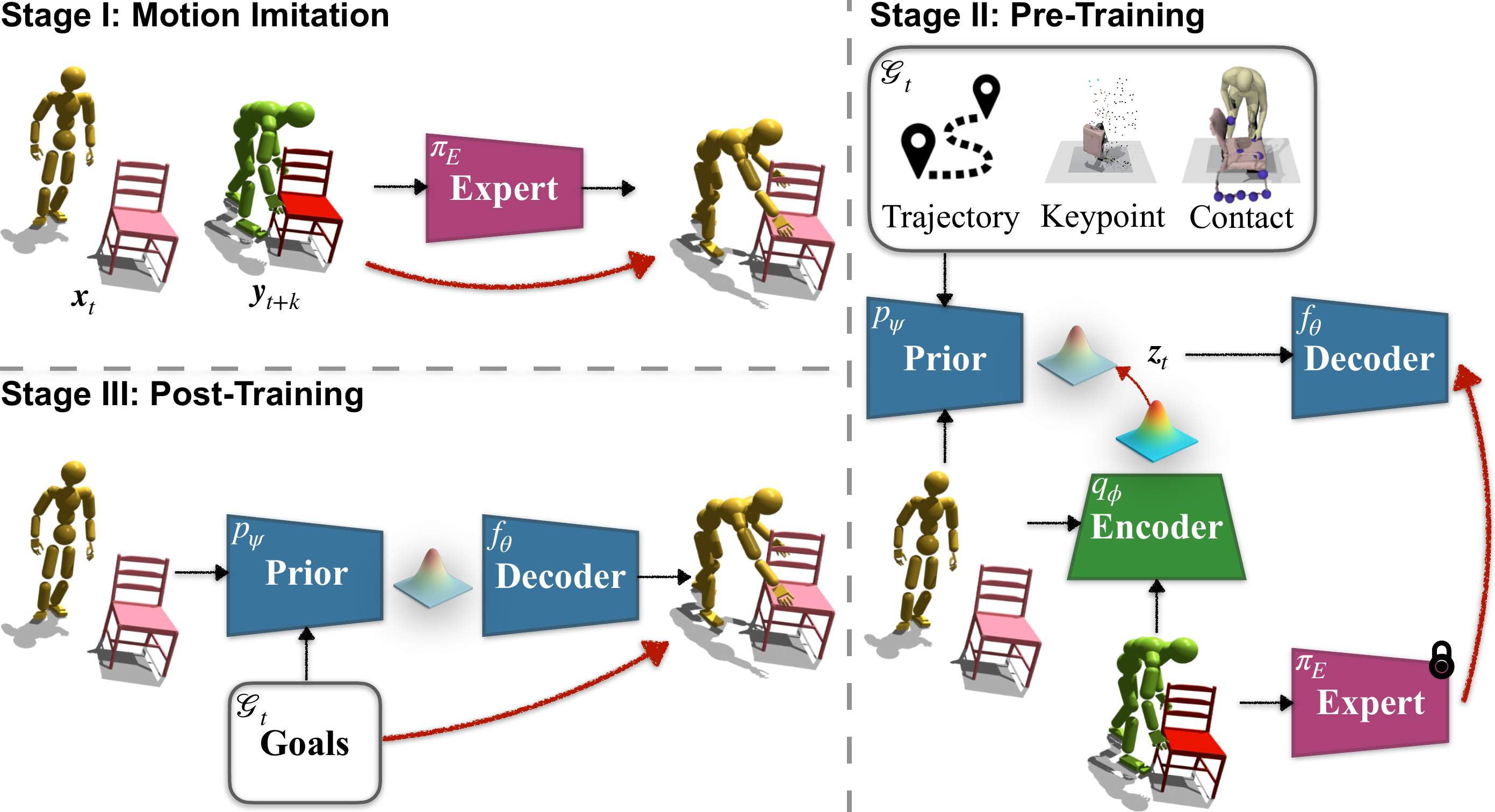}
    \caption{Overview of the proposed \ours framework. It consists of: (\textbf{I}) full-reference imitation expert training on large-scale human-object interaction data; (\textbf{II}) distillation of the expert into a variational policy with a structured latent space for skill embeddings; and (\textbf{III}) post-training of the variational policy to enhance generalization. Blue modules denote the final policy used at inference; green and red modules are training‑only components, and red arrows denote supervision signals (rewards/losses). 
    }
    \label{fig:method}
    \vspace{-0.5em}
\end{figure}

\noindent\textbf{Overview.}
Figure~\ref{fig:method} illustrates our three-stage paradigm. First, we train an expert policy \(\pi_E\) for \textit{large-scale} HOI motion imitation, incorporating data augmentation, physical perturbations, and shaped rewards to promote stable whole-body coordination and precise grasping across diverse configurations (Sec.~\ref{sec:teacher}). Second, we distill the expert into a masked conditional variational policy \(\pi\) that maps sparse goal inputs to a multi-modal distribution (Sec.~\ref{sec:distill}). Third, we finetune this policy \(\pi\) using RL to enhance robustness under unseen configurations, employing failure-state resets to encourage recovery behaviors (Sec.~\ref{sec:finetune}).
Each stage is modeled as a Markov Decision Process (MDP), which shares a consistent input formulation comprising observations and goal conditioning, as well as an output action corresponding to low-level actuation commands (Sec.~\ref{sec:state}). 

\subsection{Policy States and Actions}\label{sec:state}
\noindent\textbf{Observation.}
The policy input at time \(t\) includes an observation that aggregates human kinematics, object kinematics, and their interaction and contact states,
\(
\boldsymbol{x}_t=\big[\,\underbrace{\boldsymbol{r}^h_t,\boldsymbol{\theta}^h_t,\boldsymbol{\dot r}^h_t,\boldsymbol{\dot\theta}^h_t}_{\text{human}},\;
\underbrace{\boldsymbol{r}^o_t,\boldsymbol{\theta}^o_t,\boldsymbol{\dot r}^o_t,\boldsymbol{\dot\theta}^o_t}_{\text{object}},\;
\underbrace{\boldsymbol{D}_t,\boldsymbol{C}_t}_{\text{interaction}}\,\big].
\)
Here, the superscripts \(h\) and \(o\) denote human and object quantities, respectively. \(\boldsymbol{r}\) and \(\boldsymbol{\theta}\) denote positions and orientations, respectively; the dotted terms indicate linear and angular velocities. The interaction terms include signed distances from body segments to object surfaces \(\boldsymbol{D}_t\) and binary contacts \(\boldsymbol{C}_t\) derived from simulator contact forces, following~\cite{xu2025intermimic}. All continuous quantities are normalized in a human root‑centric and local heading frame for invariance to global placement. The human-related terms contain 52 components for the SMPL humanoid~\cite{luo2023perpetual} and 39 for the Unitree G1 robot~\cite{unitreeg1}. Each rigid body contributes one element to human-related variables in \(\boldsymbol{x}_t\), including \(\boldsymbol{D}_t\) and \(\boldsymbol{C}_t\), \eg, \(\boldsymbol{D}_t \in \mathbb{R}^{39 \times 3}\) for G1. Objects are all rigid.

\noindent\textbf{Goal Conditioning.}  
The policy is also conditioned on a set of future \emph{goals} that specify desired human-object configurations at different horizons. During training, we extract goals from reference, where each reference $\boldsymbol{y}_{t}$ shares the same state space as observation $\boldsymbol{x}_t$, including human, object, and contact components. A corresponding binary mask $\boldsymbol{m}_{t}$ indicates which components of the reference are provided to the policy~\cite{tessler2024maskedmimic}. To capture both near-term and distant intentions, we employ two types of goal conditioning: (\textbf{I}) a short-horizon preview sequence and (\textbf{II}) a long-horizon snapshot. Let $H$ denote the maximum prediction horizon, $K \subset \{1,\ldots,H\}$ a set of short-horizon offsets, and $L$ a long-horizon offset. The long-horizon offset $L$ is initialized randomly, decremented by one at each timestep, and re-sampled when it reaches zero. For each $k \in K \cup \{L\}$, we retrieve $(\boldsymbol{y}_{t+k}, \boldsymbol{m}_{t+k})$, where the mask $\boldsymbol{m}_{t+k}$ is sampled to cover every possible condition \eg, end-effector pose, object pose, human-object contacts, their combination, \etc (see Sec.~\ref{sec:mask} for details of the sampling). Each goal is represented using a \emph{masked residual encoding}:
\(
\tilde{\boldsymbol{y}}_{t+k} = \boldsymbol{m}_{t+k} \odot \Delta\!\big(\boldsymbol{y}_{t+k}, \boldsymbol{x}_t\big), \ 
\mathcal{G}_t = \{\,(\tilde{\boldsymbol{y}}_{t+k}, \boldsymbol{m}_{t+k}) \;|\; k \in K \cup \{L\}\,\},
\)
where $\odot$ denotes elementwise masking and $\Delta$ applies a log-map to rotational components and subtraction to Euclidean quantities. During inference, user-specified or model generated sparse targets can be supplied by filling only the informed components, setting the corresponding mask to one, and zeroing the rest.

\noindent\textbf{Action.}
The policy outputs an action vector \(\boldsymbol{a}_t\), defining the actuation as 
\(\boldsymbol{a}_t \in \mathbb{R}^{51\times3}\) for SMPL~\cite{loper2015smpl,MANO} and 
\(\boldsymbol{a}_t \in \mathbb{R}^{29}\) for the G1 humanoid~\cite{unitreeg1}.
Each action represents a joint position target expressed in the exponential map, 
which is subsequently converted into joint torques via proportional-derivative (PD) control. The resulting torques are applied to the corresponding joints in the physics simulator, 
driving the human-object interactions and generating the next state 
\(\boldsymbol{x}_{t+1}\) according to the simulator’s dynamics.

\subsection{InterMimic+: Full-Reference Imitation Expert}
\label{sec:teacher}

Serving as the teacher for the final policy \(\pi\), we formulate large-scale co-tracking of human and object motions following InterMimic~\cite{xu2025intermimic}. 
At each timestep \(t\), the expert policy \(\pi_E\) receives the observation along with future references, which contain complete information without masking. 
The policy outputs low-level actuation commands \(\boldsymbol{a}_t\) and is trained using Proximal Policy Optimization (PPO)~\cite{schulman2017proximal} to maximize a composite reward function:
\(
r = r_{\text{track}} \times r_{\text{energy}},
\)
where \(r_{\text{track}}\) promotes alignment between the reference \(\boldsymbol{y}_t\) and simulation state \(\boldsymbol{x}_t\), and \(r_{\text{energy}}\) encourages physically plausible and efficient behaviors. 
This formulation enforces \textit{strict adherence to the reference}.

The policy from the original InterMimic achieves high-fidelity imitation and broad loco-manipulation coverage. 
However, in practice, we observe key issues due to the policy’s strong reliance on references, which we address with our advanced version. 
(\textbf{I}) The policy shows a degradation of precision when interacting with thin or small objects, as it tends to rigidly follow reference trajectories (See Figure~\ref{fig:imit}) without utilizing fine-grained hand-object relations. 
(\textbf{II}) This limitation is more severe if the rollout deviates from reference trajectories. 
To mitigate these issues, we expand reference scope and introduce reference-free rewards.

\noindent\textbf{Expanding Reference Scope.}
To reduce reliance on reference trajectories, we apply \textit{randomization}, \textit{perturbation}, and \textit{augmentation}. We initialize each episode from reference frames with random variations in human-object poses. During rollouts, we apply sparse impulses, \ie, random velocity perturbations to the pelvis and object, to induce deviations from the references. We augment object shapes and randomize physical properties such as mass density, center-of-mass offsets, inertia, and friction, with details presented in Sec.~\ref{sec:training}. This exposes the policy to diverse dynamics, without alternating the reference. Unlike common sim-to-real practices, we do not randomize actuation parameters or add observation noise, as these do not directly enhance state or dynamics coverage. However, perturbations alone are insufficient; it is necessary to introduce a termination penalty that discourages the policy from entering failure under perturbation. We define \(r_{\mathrm{ter}} = -w_{\mathrm{ter}} \times c_{\mathrm{ter}}\), where \(c_{\mathrm{ter}}\) is triggered by a human fall or large deviations in states from references, following~\cite{xu2025intermimic}, and \(w_{\mathrm{ter}}\) is a scaling coefficient.

\noindent\textbf{Reference-Free Reward.}  
A key challenge in precise hand grasping under randomization and perturbation is that strict reference-based tracking becomes unreliable. 
To address this, we introduce a hand reward \(r_{\mathrm{h}}\) that encourages the hand to \textit{target} and \textit{wrap} around the object based on the current simulation state, rather than relying on reference trajectories. 
Details of the formulation can be found in Sec.~\ref{sec:reward}.
When combined with the reference imitation reward, it serves as a corrective term that guides the hand to orient, align, and close around the actual object, potentially deviated from the reference due to perturbations, rather than strictly following the reference trajectory. The full reward is defined as \(r_t = (r_{\mathrm{track}} \times r_{\mathrm{energy}} \times r_{\mathrm{h}}) + r_{\mathrm{ter}}\).

\subsection{InterPrior: Variational Distillation}
\label{sec:distill}

Given an imitation expert policy \(\pi_E\) (Sec.~\ref{sec:teacher}) trained to master motor skills for HOI, our objective is to distill it into a \emph{variational policy} \(\pi\). 
Unlike the expert policy \(\pi_E\), which operates under densely supervised and fully observed reference trajectories, the variational policy \(\pi\) must preserve naturalness and diversity with sparse cues. This is achieved by sampling from a latent skill distribution, which endows \(\pi\) with the capacity to generate plausible variations in action space.
Our framework builds upon~\cite{tessler2024maskedmimic,xu2025dexplore} with two new designs: (\textbf{I}) \emph{multi-modal conditioning}, including contact for versatile human-object conditioning, and (\textbf{II}) \emph{prior shaping and bounding} regularization for robustness.

\noindent\textbf{Model.}
We model the policy $\pi$ with a latent $\boldsymbol{z}_t \in \mathbb{R}^{d_z}$ to for multi-modality. As shown in Fig.~\ref{fig:method}, $\pi$ consists of:
\[
\begin{aligned}
&\textbf{Prior:} && p_\psi(\boldsymbol{z}_t \mid \boldsymbol{x}_{t-\ell:t},\mathcal{G}_t),\\
&\textbf{Encoder:} && q_\phi(\boldsymbol{z}_t \mid \boldsymbol{x}_t,\mathcal{G}_t,\boldsymbol{y}_{t:t+H}, \boldsymbol{y}_{t+L}),\\
&\textbf{Decoder:} && f_\theta(\boldsymbol{a}_t \mid \boldsymbol{x}_{t-\ell:t},\boldsymbol{z}_t).
\end{aligned}
\]
The encoder is an MLP used only during training; given the full future reference, it outputs a Gaussian $\mathcal{N}(\boldsymbol{\mu}_q,\boldsymbol{\Sigma}_q)$. In parallel, a prior Transformer encodes recent history, with history length $\ell$, and a sparse goal, producing a Gaussian $\mathcal{N}(\boldsymbol{\mu}_p,\boldsymbol{\Sigma}_p)$. Following~\cite{tessler2024maskedmimic}, we form a residual posterior \(\mathcal{N}(\boldsymbol{\mu}_p + \boldsymbol{\mu}_q, \boldsymbol{\Sigma}_q)\). During training we sample the latent skill via reparameterization:
\(
\boldsymbol{z}_t = (\boldsymbol{\mu}_p + \boldsymbol{\mu}_q) + \boldsymbol{\Sigma}_q^{1/2}\boldsymbol{\epsilon}, \ \boldsymbol{\epsilon} \sim \mathcal{N}(\mathbf{0}, \mathbf{I}),
\)
and hold $\boldsymbol{\epsilon}$ fixed within an episode to promote temporally consistency~\cite{yao2022controlvae}. During inference, only the prior is used to sample $\boldsymbol{z}_t \sim \mathcal{N}(\boldsymbol{\mu}_p,\boldsymbol{\Sigma}_p)$. The decoder MLP maps the latent and observation to the action. The decoder also includes an auxiliary head during training that reconstructs the \emph{masked} entries of the goal,
encouraging a meaningful latent space by learning to \emph{complete} intent from context.

\noindent\textbf{Bounding the Latent.}
To improve robustness and prevent unnatural behaviors induced by out-of-distribution latents, after sampling we project
\(
\boldsymbol{z}_t \leftarrow \boldsymbol{z}_t / \|\boldsymbol{z}_t\|
\)
so that the policy operates on a hypersphere, following~\cite{peng2022ase}. This simple normalization stabilizes skill learning by limiting the rare latent draws while preserving directional variability for multi-modal behaviors. Note that we apply the projection after sampling, thus KL regularization can still be computed on the Gaussian $p_\psi$ and $q_\phi$ before projection.

\noindent\textbf{Online Distillation and Regularization.}  
We utilize an online distillation framework following DAgger~\cite{ross2011reduction}, where the student policy $\pi$ learns from a mixture of expert $\pi_E$ and self-generated rollouts. Training begins with trajectories fully controlled by the expert $\pi_E$, and the ratio of student-driven states is gradually increased as learning progresses. At each step, the expert provides its action output as supervision for the student.  
The policy is optimized using a composite objective consisting of multiple loss terms:
\(
\mathcal{L}_{\text{total}}
= \mathcal{L}_{\text{ELBO}}
+ \lambda_{\text{scale}}\,\mathcal{L}_{\text{scale}}
+ \lambda_{\text{tc}}\,\mathcal{L}_{\text{tc}}.
\)
The primary objective, $\mathcal{L}_{\text{ELBO}}$, is a weighted evidence lower bound~\cite{kingma2013auto} that combines three components:  
(\textbf{I}) an \emph{imitation loss} encouraging the student to reproduce expert actions,  
(\textbf{II}) a \emph{goal reconstruction loss} promoting accurate completion of masked goal entries to align with the ground truth, and  
(\textbf{III}) a \emph{KL regularization loss} that penalizes divergence between the posterior \(\mathcal{N}(\boldsymbol{\mu}_p + \boldsymbol{\mu}_q, \boldsymbol{\Sigma}_q)\) and the prior distribution \(\mathcal{N}(\boldsymbol{\mu}_p, \boldsymbol{\Sigma}_p)\).  
We introduce two auxiliary losses to further shape the latent.
$\mathcal{L}_{\text{scale}}$ constrains the prior mean $\boldsymbol{\mu}_p$ to maintain unit magnitude, preventing degeneracy given hypersphere normalization.
$\mathcal{L}_{\text{tc}}$ encourage consecutive prior distributions to remain similar across time steps.  
Details of these losses are provided in Sec.~\ref{sec:reward}.

\subsection{InterPrior: Post-Training Beyond Reference}\label{sec:finetune}

The distilled policy $\pi$ (Sec.~\ref{sec:distill}) exhibits goal following, yet it is brittle when the goal or human-object state drifts off the dataset distribution, \eg, during transitions between skills. Unlike human-only motion~\cite{luo2023universal} or small-object grasping~\cite{tessler2025maskedmanipulator}, loco‑manipulation tasks with coupled affordances span a far larger configuration space that references alone cannot cover. This follows from the learning dynamics of distillation: training proceeds by replaying dataset trajectories. 
Our key observation is that the pretrained $\pi$ provides a strong and natural initialization for RL finetuning as a local optimizer that expands its scope along three axes: (\textbf{I}) recover from near‑failure or failure states, (\textbf{II}) explore unseen yet plausible configurations without trajectory replay, and at the same time (\textbf{III}) preserve the naturalness of behaviors encoded by the pretrained policy.
A natural alternative is to sample novel multi-frame trajectories that combine diverse human, object, and contact configurations and then train the policy to track them~\cite{luo2024grasping}, but this requires a strong trajectory sampler, which is particularly challenging at loco-manipulation scale. Instead, we target \emph{single-frame} goals: composing goals observed in data can induce unseen configurations, and we further combine such goals with randomized initializations and offsets to systematically broaden the state distribution encountered during RL.

\noindent\textbf{In-Betweening for Finetuning.}
To mitigate the cost of exhaustive trajectory sampling, we formulate finetuning as an \emph{in-betweening} task, where the policy tracks from a randomly sampled initial configuration toward a single-frame goal randomly drawn from the dataset. The policy is rewarded for progressing toward this sampled goal.
The reward is defined as,
\[
\begin{aligned}
r^{\mathrm{PT}}_t
&=\big(r_{\text{energy}} \times r_{\mathrm{h}}\big)\;+\;r_{\text{goal}}\;+\;r_{\mathrm{ter}},\\
r_{\text{goal}}
&= 
\begin{cases}
r_{\text{succ}}, & \text{if }\big\|{\boldsymbol{m}}_{t+L}\odot\Delta(\tilde{\boldsymbol{y}}_{t+L},\,\boldsymbol{x}_t)\big\|_1 < \tau,\\[3pt]
0, & \text{otherwise.}
\end{cases}
\end{aligned}
\]
where the terms $r_{\text{energy}}$, $r_{\mathrm{ter}}$, and $r_{\mathrm{h}}$ are defined in Sec.~\ref{sec:teacher}. Since the goal is arbitrary by the random masking, we do not use a dense distance-based reward. The goal reward $r_{\text{goal}}$ provides a sparse success signal that activates when the masked feature distance between the current state $\boldsymbol{x}_t$ and target $\tilde{\boldsymbol{y}}_{t+L}$ falls below a threshold $\tau$. \(r_{\text{succ}}\) is a constant. 

\noindent\textbf{Learning New Skills.}
As shown in Figure~\ref{fig:teaser}, our RL finetuning can expand the distilled policy by handling two common regimes. (\textbf{I}) \emph{In-distribution extensions} reuse and compose behaviors already supported by the demonstrations. A representative example is \emph{regrasping}, which arises naturally from goal-conditioned in-betweening: training the policy to reach goals from diverse initializations and perturbed states encourages self-correction from near-failure outcomes without additional supervision. (\textbf{II}) \emph{Out-of-distribution skills} must be learned explicitly when the required behavior is absent from the dataset. A representative example is \emph{getting up}. Following prior practice~\cite{vainshtein2025task,pan2025tokenhsi}, we append a learnable \emph{token} to the (Sec.~\ref{sec:distill}) to indicate this new subtask and add an auxiliary reward that encourages upright posture and center-of-mass elevation (Sec.~\ref{sec:reward}). 

\noindent\textbf{Prior Preservation.}
During finetuning, rather than freezing network components to mitigate catastrophic forgetting as in prior work~\cite{pan2025tokenhsi,vainshtein2025task}, we adopt a simple multi-objective schedule. Specifically, we maintain a subset of environments that continue optimizing the original distillation objective (Sec.~\ref{sec:distill}), while the remaining environments perform RL finetuning (Sec.~\ref{sec:reward}). This anchors the policy to the pretrained prior during adaptation without restricting model capacity. Given the environment mixtures and the joint execution of RL and distillation, we distribute tasks across multiple GPUs and aggregate gradients via a map-reduce scheme. 
Further details are provided in Sec.~\ref{sec:reward}.

\section{Experiments}
\label{sec:exp}
We evaluate \ours on two tasks: (\textbf{I}) \textit{full-reference tracking} and (\textbf{II}) \textit{sparse goal following}. The evaluation covers snapshot, trajectory, and contact specification, as well as their \textit{compositions}. Since our goal representation is formed by masking arbitrary subsets of targets, these settings subsume a \textit{broad family of task formulations}, ranging from single-frame constraints to multi-step trajectories over different joints and contacts. We further study \ours as a reusable prior for novel objects, and for tracking trajectories generated by kinematic models (Sec.~\ref{sec:add_exp}).

\noindent\textbf{Datasets.}
We employ the InterAct~\cite{xu2025interact} dataset with its preprocessing, which features diverse daily interactions encompassing a wide range of subjects and objects. Following~\cite{xu2025intermimic}, we use the OMOMO subset~\cite{li2023object} repaired by their teacher rollout.
To assess generalizability, we apply \ours to other InterAct subsets including selected data from BEHAVE~\cite{bhatnagar22behave} and HODome~\cite{zhang2023neuraldome}. We exclude interactions dominated by soft-body dynamics (\eg, backpack shoulder straps) when choosing evaluation examples.

\noindent\textbf{Baselines and Tasks.} We focus on baselines that cover diverse objects and skills and therefore omit methods that are for single object or task-specific proficiency~\cite{wang2023physhoi,yu2025skillmimic,pan2025tokenhsi}.
(\textbf{I}) \emph{Full-reference tracking.} We compare against the original InterMimic~\cite{xu2025intermimic}, with \ours, which supports full-reference imitation by removing masks. Evaluations target challenging regimes involving \emph{thin-object interactions} and \textit{initialization noise}.
(\textbf{II}) \emph{Sparse goal following.} We evaluate the complete \ours framework against adapted MaskedMimic~\cite{tessler2024maskedmimic,tessler2025maskedmanipulator}, to our task under identical goals, following Figure~\ref{fig:teaser}:
(a) \textit{Snapshot goals:} a ground truth frame specifies a few human joints or object position in the long term;
(b) \textit{Trajectory goals:} a sequence of ground-truth keyframes defines the a few joints or object trajectories;
(c) \textit{Contact goals:} a contact schedule specifies the desired active contact regions on objects, which will be converted to goals for human joints;
(d) \textit{Multi-goal chaining:} To evaluate long-horizon robustness, we concatenate three randomly sampled ground-truth subgoals, each canonicalized with respect to the preceding one. The concatenated sequence may include a mixture of snapshot, trajectory, and contact-following segments, with randomized goal transitions. For consistency, the same goals are used across all baselines;
(e) \textit{Random initialization:} To test motion coverage, we initialize the humanoid within five meters of the object and define the task as lifting the object by 0.5 meters from its initial position.

\noindent\textbf{Metrics.}
(\textbf{I}) \emph{Full-reference tracking.}
Following~\cite{xu2025intermimic}, we report the following metrics:
(a) \textit{Success Rate (SR)}: the proportion of rollouts completed without violating the early-termination criteria;
(b) \textit{Human Position Error} $E_{\text{h}}$ (m): the mean per-joint positional deviation between the simulated and reference humans, excluding hands due to the missing ground truth from the dataset; and
(c) \textit{Object Position Error} $E_{\text{o}}$ (m): the mean positional deviation between the simulated and reference objects.
(\textbf{II}) \emph{Sparse goal following.}
The evaluation metrics include:
(a) \textit{Success Rate (SR)}; (b) \textit{Human} and \textit{Object Errors} ($E_{\text{h}}$, $E_{\text{o}}$): the deviation from the target goal state, computed over the unmasked region; and
(c) \textit{Failure Rate (Fail)}: proportion of rollouts that directly fail \eg, fall.
More details are presented in Sec.~\ref{sec:add_exp}.

\noindent\textbf{Implementation Details.}
All control policies operate at 30\,Hz in IsaacGym~\cite{makoviychuk2021isaac}.
The imitation expert policy, along with the encoder and decoder used during distillation, are implemented as MLPs with hidden layers of (1024, 1024, 512).
The prior network is a four-layer Transformer encoder, and the critics use the same MLP architecture for expert training and RL finetuning.
We retrain InterPrior on the G1 embodiment using our three-stage paradigm. During the first stage, we incorporate additional rewards and domain randomization to enhance stability on G1 and facilitate robust sim-to-sim transfer. All auxiliary rewards are multiplied with the imitation reward in exponential form $\exp(-\,\cdot)$, except for the termination term, which is added directly.  The formulation of each G1-specific reward term is provided in Table~\ref{tab:reward_stage1}, and the dynamics randomization ranges used during training are summarized in Table~\ref{table:dr}. We exclude thin-geometry objects for G1 because we do not include dexterous hands supporting single-hand grasps.

\begin{figure}
  \centering
  \includegraphics[width=\columnwidth]{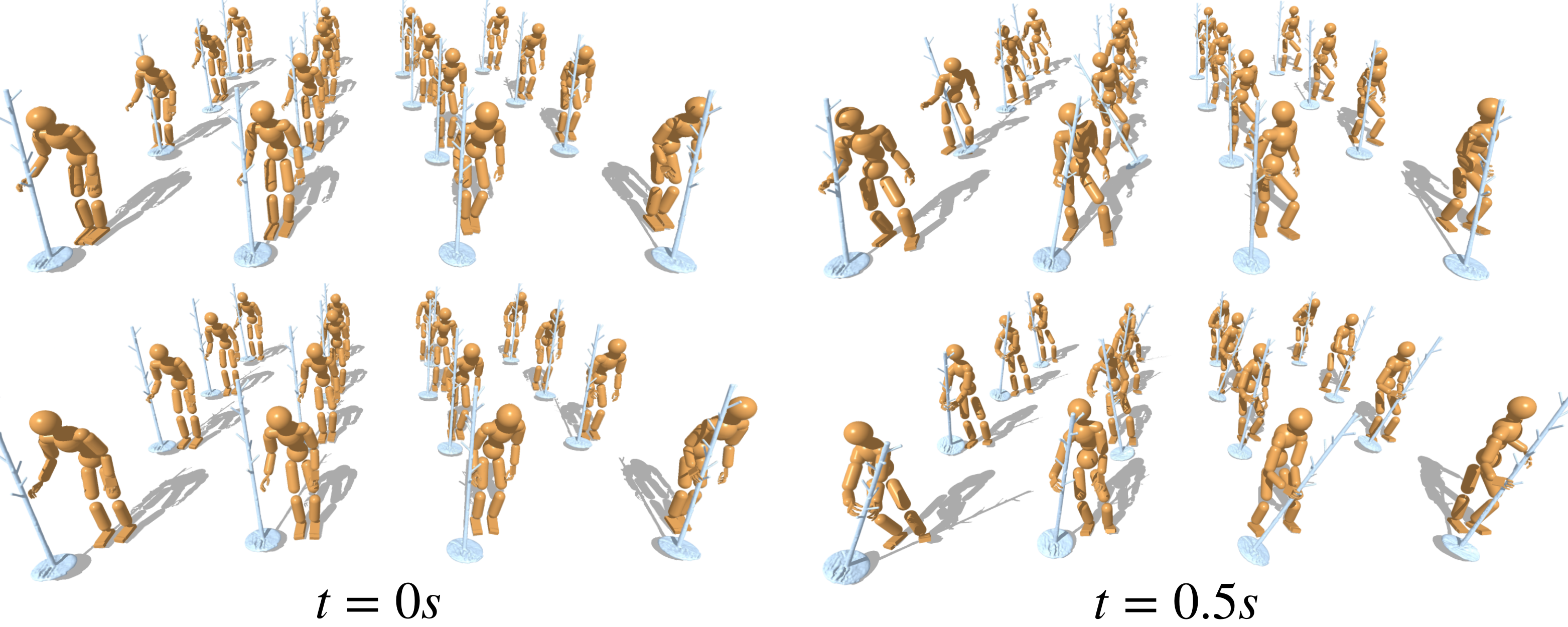}
  \caption{\textbf{Qualitative comparison} of same reference imitation between InterMimic~\cite{xu2025intermimic} (\textbf{top}) and our InterMimic+ (\textbf{bottom}). InterMimic strictly follows the reference humanoid motion but fails to grasp the thin cloth stand when initialized with perturbations.}
  \label{fig:imit}
  \vspace{-0.5em}
\end{figure}

\begin{figure}
  \centering
  \includegraphics[width=\columnwidth]{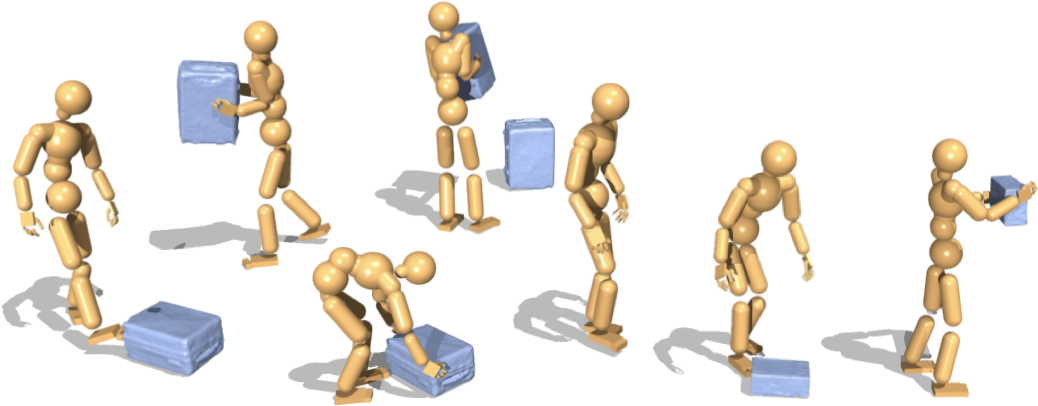}
  \caption{\textbf{Qualitative results} on a multi-object task. The model input is shifted to the second object once the first object is released.}
  \label{fig:long}
  \vspace{-0.5em}
\end{figure}

\subsection{Quantitative Results}
(\textbf{I}) \emph{Full-reference tracking.}
Table~\ref{tab:cross_tracking} shows that \ours{} achieves higher success rates under thin-geometry interactions and initialization noise. While InterMimic attains lower position error by strictly tracking the reference, \ours{} sometimes yields slightly higher human position error because it intentionally deviates when needed to re-align contact, trading strict tracking for interaction completion.
(\textbf{II}) \emph{Goal-conditioned tasks.}
Under identical goal specifications (Table~\ref{tab:goal_conditioned}), \ours{} consistently improves success and reduces errors, with the largest gains on long-horizon multi-goal chaining and random-initialization stress tests. Distillation-based policies (including \ours{} pre-RL) fit the demonstration-induced state distribution; long rollouts with goal switching can enter under-covered intermediate states, causing drift and failure. RL finetuning directly trains the policy to reach sparse targets from diverse initializations, improving interpolation across goal sequences and recovery from off-distribution states. The position error trends follow a goal-sparsity continuum: broader state coverage benefits sparse goals more, and the gap narrows as goals densify. With full-reference tracking (Table~\ref{tab:cross_tracking}), InterMimic for strict tracking achieves the lowest errors.

\subsection{Qualitative Results}
\begin{figure}
  \centering
  \includegraphics[width=\columnwidth]{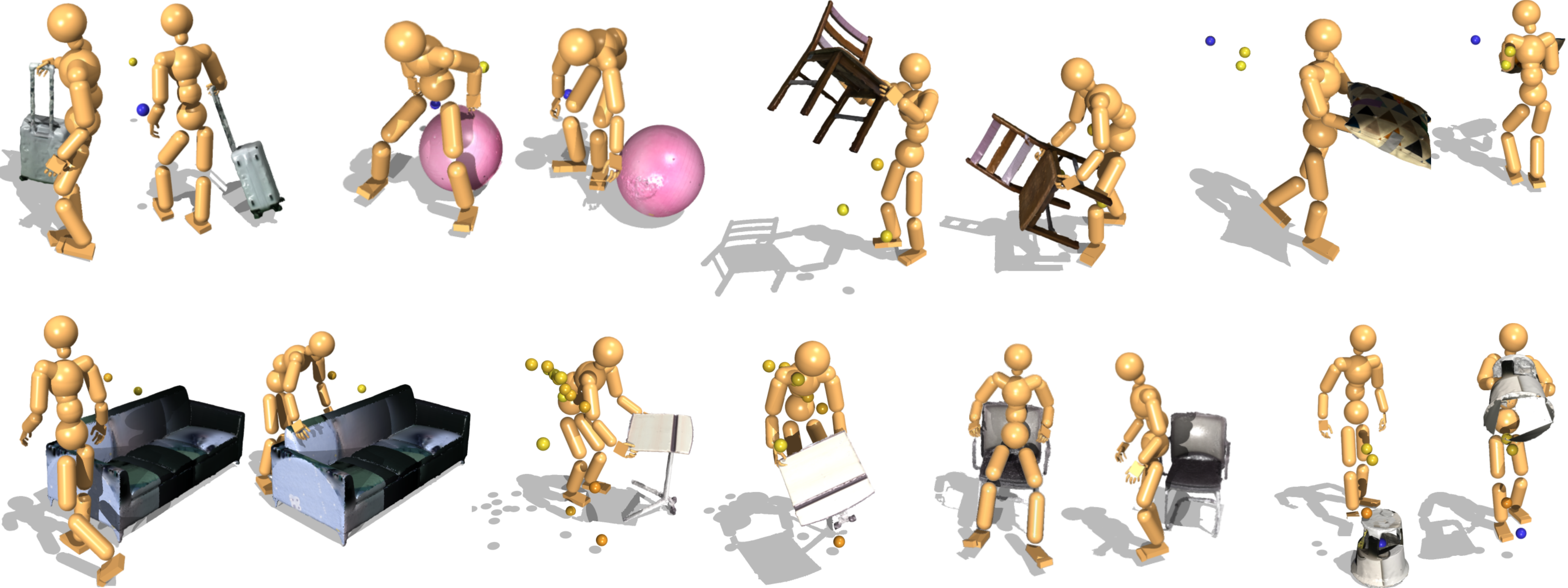}
  \caption{\textbf{Zero-shot qualitative results.} A single \ours{} model trained from OMOMO~\cite{li2023object} demonstrates generalization to \textit{unseen} objects and interactions from BEHAVE~\cite{bhatnagar22behave} and HODome~\cite{zhang2023neuraldome}.}
  \label{fig:adapt}
  \vspace{-0.5em}
\end{figure}

\begin{figure}
  \centering
  \includegraphics[width=\columnwidth]{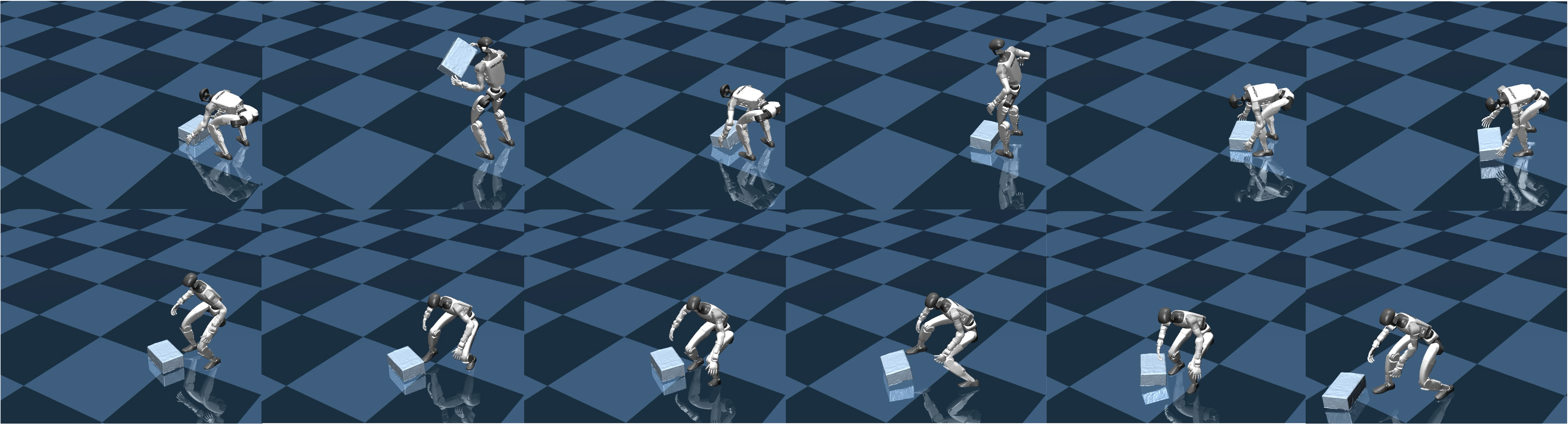}
  \caption{\textbf{Qualitative results} on sim-to-sim from IsaacGym~\cite{makoviychuk2021isaac} to MuJoCo~\cite{todorov2012mujoco} with object trajectory as condition, showing a sustained interaction involving box pickup, pushing, and kicking.}
  \label{fig:sim2sim}
  \vspace{-0.5em}
\end{figure}

\begin{figure*}
  \centering
  \includegraphics[width=\textwidth]{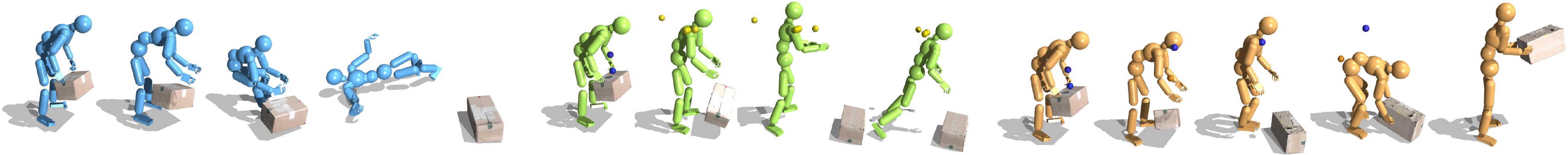}
    \caption{\textbf{Qualitative comparison} between InterMimic~\cite{xu2025intermimic} (\textbf{left}, full reference), MaskedMimic~\cite{tessler2024maskedmimic} (\textbf{middle}), and our \ours (\textbf{right}) on unseen and imperfect interactions from the BEHAVE~\cite{bhatnagar22behave} dataset. \ours can recover from data imperfection and continue the rollout.} 
  \label{fig:compare}
  \vspace{-0.5em}
\end{figure*}

(\textbf{I}) \textit{Full-reference tracking.}
Figure~\ref{fig:imit} shows that InterMimic rigidly follows the reference but often fails to acquire or maintain contact on thin geometries under perturbations. In contrast, our tracking policy allows small, targeted deviations to correct hand-object alignment, producing stable grasps and more reliable completion.
(\textbf{II}) \textit{Long-horizon tasks.}
Figures~\ref{fig:long} and \ref{fig:teaser} show that \ours{} sustains minute-long whole-body interaction with multiple objects and smooth transitions across skills (\eg, approach, grasp, lift, reposition). When drift begins (contact or balance), \ours{} self-corrects instead of compounding errors, consistent with the robustness induced by RL finetuning.
(\textbf{III}) \textit{Novel objects and interactions.}
Figures~\ref{fig:adapt} and \ref{fig:compare} demonstrate zero-shot generalization to unseen objects and interaction styles. Guided only by sparse snapshot goals, \ours{} complete unspecified degrees of freedom and converge to feasible contact, even the original data in BEHAVE~\cite{bhatnagar22behave} and HODome~\cite{zhang2023neuraldome} is for different human shape.
(\textbf{IV}) \textit{Sim-to-sim transfer.}
Figure~\ref{fig:sim2sim} illustrates transfer from IsaacGym~\cite{makoviychuk2021isaac} to MuJoCo~\cite{todorov2012mujoco}: \ours{} maintains coherent long-horizon interactions under object-conditioned goals, showing the potential to transfer to the real world.

\begin{table*}
\centering
\setlength{\tabcolsep}{3.2pt}
\renewcommand{\arraystretch}{1.1}
\caption{\textbf{Quantitative evaluation} and \textbf{ablation study} on in-distribution goal-conditioned tasks, including snapshot, trajectory, contact (Figure~\ref{fig:teaser}), plus out-of-distribution \textbf{stress tests} on challenging scenerio, such as long-horizon multi-goal chains and object lifting under random human initialization. For the random initialization, only the object is assigned a goal, thus the human error is omitted.}
\label{tab:goal_conditioned}
\resizebox{\textwidth}{!}{%
\begin{tabular}{ll|cccc|cccc|cccc|ccc|cc}
\toprule
\multicolumn{2}{c|}{Method} 
& \multicolumn{4}{c|}{Snapshot} 
& \multicolumn{4}{c|}{Trajectory} 
& \multicolumn{4}{c|}{Contact} 
& \multicolumn{3}{c|}{\textbf{Chain}} 
& \multicolumn{2}{c}{\textbf{Rand Init}} \\
\cmidrule(lr){1-2}\cmidrule(lr){3-6}\cmidrule(lr){7-10}\cmidrule(lr){11-14}\cmidrule(lr){15-17}\cmidrule(lr){18-19}
Variant & Additions (cumulative) 
& Succ $\uparrow$ & $E_{\text{h}}$ $\downarrow$ & $E_{\text{o}}$ $\downarrow$ & Fail $\downarrow$
& Succ $\uparrow$ & $E_{\text{h}}$ $\downarrow$ & $E_{\text{o}}$ $\downarrow$ & Fail $\downarrow$
& Succ $\uparrow$ & $E_{\text{c}}$ $\downarrow$ & $E_{\text{o}}$ $\downarrow$ & Fail $\downarrow$
& Succ $\uparrow$ & $E_{\text{h}}$ $\downarrow$ &  $E_{\text{o}}$ $\downarrow$
& Succ $\uparrow$ & $E_{\text{o}}$ $\downarrow$ \\
\midrule
MaskedMimic~\cite{tessler2024maskedmimic} & InterMimic~\cite{xu2025intermimic} as Expert 
& 64.2 & 29.3 & 22.1 & 12.6  & 88.0 & 9.0 & 8.1 & 8.5  & 52.2 & 49.2 & 25.7 & 13.9  & 29.1 & 40.2 & 43.9 & 31.7 & 26.8 \\
\midrule
\multirow{6}{*}{\ours{} (\textbf{Ours})}
& InterMimic+ as Expert 
& 71.4 & 18.6 & 11.7 & 11.0  & 92.7 & 8.2 & 7.7 & 5.2  & 69.3 & 25.6 & 18.2 & 9.7  & 33.9 & 37.1 & 39.6   & 30.1 & 22.1  \\
& + Latent Shaping Loss
& 74.9 & 20.4 & 15.5 & 10.6 & 92.4 & \textbf{7.9} & \textbf{6.6} & 5.3  & 71.9 & 26.7 & 15.3 & 11.9  & 40.0 & 37.0 & 40.8  & 30.9 & 13.9  \\
& + Bounded Latent \& Observations 
& 89.1 & \textbf{11.7} & \textbf{8.9} & 6.0  & 93.6 & 8.1 & \textbf{6.6} & 4.6  & 88.5 & 17.0 & \textbf{8.1} & 5.4  & 45.1 & 31.5 & 37.2  & 41.1 & 19.6  \\
& + RL Finetuning \, (= \textbf{full}) 
& \textbf{90.0}  & 13.6 & 9.5 & \textbf{3.7}  & \textbf{94.6} & \textbf{7.9} & 6.9 & \textbf{2.5}  & \textbf{90.7} & \textbf{15.9} & 9.9 & \textbf{2.9}  & \textbf{68.8} & \textbf{30.2} & \textbf{35.7}  & \textbf{88.6} & \textbf{11.9} \\
\bottomrule
\end{tabular}%
}
\vspace{-0.5em}
\end{table*}

\begin{table}
\centering
\setlength{\tabcolsep}{5.5pt}
\renewcommand{\arraystretch}{1.1}
\caption{\textbf{Quantitative evaluation} of full-reference imitation on OMOMO with \textit{thin objects} and \textit{initialization perturbations}, and adaptation to \textit{novel object} and \textit{interaction} skills, evaluated before and after finetuning on new data. For novel interactions, $E_{\text{h}}$ and $E_{\text{o}}$ not directly comparable since \ours now uses random sparse goals. Results show that InterPrior functions as a \textbf{reusable prior} with stronger adaptation capability than the full-reference imitator.}
\label{tab:cross_tracking}
\resizebox{\columnwidth}{!}{%
\begin{tabular}{l|ccc|c|c}
\toprule
 & \multicolumn{3}{c|}{OMOMO~\cite{li2023object} select} 
 & \multicolumn{1}{c|}{BEHAVE~\cite{bhatnagar22behave}}
 & \multicolumn{1}{c}{HODome~\cite{zhang2023neuraldome}} \\
\cmidrule(lr){2-4}\cmidrule(lr){5-5}\cmidrule(lr){6-6}
Method 
& SR $\uparrow$ & $E_{\text{h}}$ $\downarrow$ & $E_{\text{o}}$ $\downarrow$
& SR $\uparrow$
& SR $\uparrow$ \\
\midrule
InterMimic~\cite{xu2025intermimic} 
& 63.9 & \textbf{7.1} & \textbf{11.4} 
& 10.7
& 27.8 \\
InterMimic + finetuning 
& / & / & /
& 38.9
& 55.5 \\
\ours{} 
& \textbf{83.2} & 8.9 & 11.7 
& 27.4
& 40.1 \\
\ours{} + finetuning 
& / & / & /
& \textbf{52.0}
& \textbf{72.4} \\
\bottomrule
\end{tabular}%
}
\vspace{-0.5em}
\end{table}

\subsection{Ablation Study}
We conduct a cumulative ablation study reported in Table~\ref{tab:goal_conditioned}. Starting from a MaskedMimic baseline with an InterMimic expert, we progressively enable the components of \ours: upgrading to an InterMimic+ expert, incorporating the latent shaping loss, bounding both latent and observation spaces, and finally applying RL finetuning.

\noindent\textbf{Impact of Latent Shaping and Bounding.} Introducing the latent shaping loss yields modest improvements on in-distribution tasks but provides clear gains for long-horizon behavior and under random initialization. This indicates that a well-shaped and properly bounded latent is essential for mitigating drift in challenging, contact-rich interactions.

\noindent\textbf{Effectiveness of Finetuning.} Comparing the full \ours model with the variant before finetuning shows that RL finetuning chiefly enhances robustness. The improvement is also more pronounced on stress tests, suggesting that finetuning helps the policy exploring the feasible motion space and recover from distributional shift, while maintaining the policy with similar precision on standard tasks.

\noindent\textbf{Impact of Finetuning on Trajectory Following.}
As discussed in Sec.~\ref{sec:finetune}, our \textit{in-betweening} finetuning is applied only on snapshot goals rather than full trajectories, which may raise concerns about degrading trajectory-following performance. However, as shown in Table~\ref{tab:goal_conditioned}, trajectory following is well preserved for two reasons: (\textbf{I}) the finetuning procedure does not alter the model under trajectory-conditioned inputs, which are explicitly protected by a concurrent distillation loss; and (\textbf{II}) we redefine a snapshot goal if deviations from the target trajectory appears, and thus trajectory-following can implicitly benefit from the RL finetuning on snapshot goal following.

\noindent\textbf{Scalable Prior.} Beyond the generalization results in Figure~\ref{fig:adapt}, Table~\ref{tab:cross_tracking} and Figure~\ref{fig:compare} further demonstrate that \ours scales more robustly to novel objects and interactions, with or without finetuning, compared to the full-reference InterMimic baseline. A key factor is the prevalent dataset imperfections. For example, in Figure~\ref{fig:compare}, baselines fail as contact artifacts cause failure initialization, whereas \ours can re-establish contact and continue the task. This flexibility allows the learned model to better absorb additional interaction data, even when such data are imperfect.

\noindent\textbf{Failure Cases.}
Despite its improved robustness over the baselines, InterPrior still exhibits failure modes, as shown in Figure~\ref{fig:failure}. The human loses contact and moves without the object, whereas the baseline demonstrates a significantly higher failure rate, often resulting in human fall. We find typical failure scenarios include: (\textbf{I}) challenges with extremely thin or elongated objects that were unseen during training; and (\textbf{II}) partial goal completion in multi-goal chaining, where canonicalization introduces large alignment discrepancies, leading the policy to favor maintaining balance over achieving precise goal configurations.

\section{Conclusion} 
We present \ours, a physics-based generative motion controller that scales human-object interaction by combining large-scale imitation distillation with reinforcement finetuning. Using a distilled, goal-conditioned latent policy and optimizing it with RL yields a controller that maintains natural whole-body coordination while substantially improving robustness and competence. It composes loco-manipulation skills, transitions smoothly, and recovers from failures across diverse contact and dynamic conditions. This decoupled recipe broadens task, skill, and dynamics coverage while enabling interactive control and can be applied to different embodiments. We hope this scalable paradigm to provide a practical recipe for humanoid loco-manipulation. Future directions include integrating perception, language-conditioned goals, and richer affordances to advance \ours toward robust sim-to-real assistive manipulation and teleoperation.

{
    \small
    \bibliographystyle{ieeenat_fullname}
    \bibliography{main}
}

\clearpage
\setcounter{page}{1}
\maketitlesupplementary

\setcounter{table}{0}
\renewcommand{\thetable}{\Alph{table}}
\renewcommand*{\theHtable}{\thetable}
\setcounter{figure}{0}
\renewcommand{\thefigure}{\Alph{figure}}
\renewcommand*{\theHfigure}{\thefigure}
\setcounter{section}{0}
\renewcommand{\thesection}{\Alph{section}}
\renewcommand*{\theHsection}{\thesection}

\noindent In this supplementary, we provide additional details of our InterPrior framework with extended experiments:

\begin{enumerate}[label=(\textbf{\roman*})]
    \item Sec.~\ref{sec:demo} describes the organization of the demo video.
    \item Sec.~\ref{sec:simulation} details the overall simulation configuration.
    \item Sec.~\ref{sec:mask} provides additional information on our goal representation, \eg, how snapshot, trajectory, and contact goals are constructed at training and evaluation time with the masks.
    \item Sec.~\ref{sec:reward} gives a comprehensive explanation on: (\textbf{I}) the detailed formulation of the reference-free hand reward; (\textbf{II}) the losses used for variational distillation and latent shaping, and (\textbf{III}) RL finetuning.
    \item Sec.~\ref{sec:training} specifies additional implementation details, including network architectures, training schedules, and how we apply data augmentation to expert training, as well as additional techniques we use during G1 training for sim-to-sim experiments.
    \item Sec.~\ref{sec:add_exp} presents further qualitative results, \eg, the integration of InterPrior with kinematic HOI generators, additional details of metrics, and failure cases.
    \item Sec.~\ref{sec:discuss} examines the limitations of our current system and its potential societal implications.
\end{enumerate}

\etocdepthtag.toc{mtappendix}
\etocsettagdepth{mtchapter}{none}
\etocsettagdepth{mtappendix}{subsection}
{
  \hypersetup{
    linkcolor = black
  }
  \tableofcontents
}

\section{Demo Video} \label{sec:demo}

The demo video on the \href{https://sirui-xu.github.io/InterPrior/}{webpage} visualizes behaviors produced by InterPrior across settings detailed in the following. All sequences are rendered from the physics simulator~\cite{makoviychuk2021isaac,todorov2012mujoco} using the same SMPL~\cite{loper2015smpl,MANO} and G1~\cite{unitreeg1} model as for training. No post-processing is applied other than camera selection and cropping for visualization.

\noindent\textbf{Core Capability.} We show examples of snapshot, trajectory, and contact-conditioned control corresponding to the scenarios illustrated in Figure~\ref{fig:teaser} of the main paper, for objects with diverse shapes.

\noindent\textbf{Failure Recovery and Regrasping.} We visualize rollouts perturbed or initialized from failure states. The video highlights re-approaching, re-grasping, and recovery from falls as described in Sec.~\ref{sec:finetune}.

\noindent\textbf{Long-Horizon Multi-Goal Chains.} We include long sequences where three canonicalized sub-goals are chained (Sec.~\ref{sec:exp}, ``Chain'' tasks) and the policy must transition smoothly between different interaction while maintaining task success.

\noindent\textbf{Diverse Task Execution from the Same Goal.} We show that our model is able to control the simulated human achieving the same task with different execution.

\noindent\textbf{Baseline Comparison.} We demonstrate that InterPrior achieves superior performance compared to existing baseline methods~\cite{tessler2024maskedmimic,tessler2025maskedmanipulator,xu2025intermimic}.

\noindent\textbf{Novel Interaction Generalization.} We visualize qualitative results on BEHAVE~\cite{bhatnagar22behave} and HODome~\cite{zhang2023neuraldome}, as a complementary to Figure~\ref{fig:adapt} and Figure~\ref{fig:compare} in the main paper.

\noindent\textbf{Interaction with multiple objects.} We showcase that InterPrior supports human interactions with multiple objects, without requiring any task-specific training.

\noindent\textbf{Sim-to-Sim for G1.} We include more examples of the G1 humanoid with sim-to-sim transfer, as a complementary to Figure~\ref{fig:sim2sim}, for controlling a humanoid only based on object future snapshot goal.

\noindent\textbf{Interactive Steering Control.} Finally, we show real-time keyboard control where a user steers high-level goals and InterPrior produces coherent whole-body motion online.

\section{Simulation} \label{sec:simulation}

All experiments are performed in IsaacGym~\cite{makoviychuk2021isaac} with the GPU PhysX backend. Control policies run at {30}{Hz}, while the simulator is stepped at {60}{Hz} with two internal substeps per control step. The main simulation hyperparameters are summarized in Table~\ref{tab:physics_hyper}.

\begin{table}
  \centering
  \caption{Simulation hyperparameters used in IsaacGym~\cite{makoviychuk2021isaac}. We largely follow the settings from prior work~\cite{xu2025intermimic,wang2023physhoi}.}
  \begin{tabular}{l|l}
    \toprule
    Hyperparameter & Value \\
    \midrule
    Simulation step $\Delta t$ & $1/60\,\text{s}$\\
    Control step $\Delta t$ & $1/30\,\text{s}$\\
    \midrule
    Physics substeps per control step & 2 \\
    Position solver iterations & 4 \\
    Velocity solver iterations & 1 \\
    Contact offset & 0.02 \\
    Rest offset & 0.0 \\
    Max depenetration velocity & 100 \\
    \midrule
    Object \& ground restitution & 0.7 \\
    Object \& ground friction & 0.9 \\ 
    Object density & 200 \\
    Max convex hulls per object & 64 \\
    Object rest offset & 0.01 \\
    \bottomrule
  \end{tabular}
  \label{tab:physics_hyper}
\end{table}

We introduce a small object rest offset to reduce human-object interpenetration, especially for thin geometries. Although this slightly enlarges the effective collision boundary, it avoids the substantial cost associated with increasing solver accuracy to compensate for collision handling.

\section{Goal Formulation} \label{sec:mask}

This section details the construction of snapshot, trajectory, and contact goals and the associated masks used. Specifically, a goal state $\boldsymbol y_t$ shares the same structure as the observation $\boldsymbol x_t$, and a binary mask $\boldsymbol m_t$ indicates which components of $\boldsymbol y_t$ are provided to the policy.

\subsection{Horizon for Goals}

\noindent\textbf{Short-Horizon Preview.}  
We use a small set of offsets $K=\{1, 2, 4, 16\}$ to provide short-horizon previews relative to the current timestep $t$. For each offset $k \in K$, we construct a goal pair $(\boldsymbol y_{t+k}, \boldsymbol m_{t+k})$.

\noindent\textbf{Long-Horizon Snapshot.}  
A long-horizon offset sampled by $L \in [1, 128]$ defines a single far-future goal $(\boldsymbol y_{t+L}, \boldsymbol m_{t+L})$. During training, $L$ is initialized randomly at the start of each episode and then decremented each timestep, being resampled once it reaches zero. Although termed a long-horizon snapshot, its value naturally decreases at each step and may temporarily fall below the short-horizon offsets.

\subsection{Stochastic Mask Sampling during Training}

During training, masks are not tied to specific tasks (snapshot; trajectory; contact).
Instead, we randomly decide which parts of the future state are revealed to the policy, so that the policy is exposed to a \textit{wide variety} of partial and sparse goals, following~\cite{tessler2024maskedmimic}.
We operate at the level of rigid bodies, including objects with following three rules:

\noindent\textbf{Body-Wise Masking.}
Visibility is enforced at the body level. For each rigid body, we maintain a single binary variable. If it is \emph{false}, all all state features associated with that body at time $t{+}k$ are masked out, positions, orientations, and linear and angular velocities. The same rule applies to the entries in the interaction vectors $D_{t+k}$ and the contact state $C_{t+k}$, defined in Sect.~\ref{sec:state}, which are masked or revealed together.
    
\noindent\textbf{Independent Sampling in Rigid Bodies.}
At each horizon offset $k$, each body is sampled independently according to a fixed Bernoulli distribution: human-state and interaction components are revealed with probability $0.1$, and object components with probability $0.5$. This procedure produces diverse, randomly constructed combinations of visible and masked human, object, and contact features, rather than relying on any task-specific mask templates.
    
\noindent\textbf{Temporal Consistency of Masks.}
    To avoid flickering visibility, masks evolve over time with a high probability of staying the same and a small probability of being re-sampled.
    Concretely, for $k > 1$ we define a first-order Markov process:
    \[
        \boldsymbol m_{t+k} =
        \begin{cases}
            \boldsymbol m_{t+k-1}, & \text{with probability } 1 - p_{\text{reset}},\\
            \text{Bernoulli}(\boldsymbol p_{\text{vis}}), & \text{with probability } p_{\text{reset}}.
        \end{cases}
    \]
    Here $p_{\text{reset}} = {0.01}$ ensures that once a body is masked or unmasked, it tends to remain in that state for multiple steps, while occasional resets still diversify the masks. The visibility probabilities $\boldsymbol p_{\text{vis}}$ follow the design above.
    
\subsection{Task Definition for Inference}

During inference, masks are constructed according to the target task. For a given task, the visibility pattern remains fixed throughout the rollout. The only exception is the multi-goal chaining setting, where we resample a new mask whenever the controller transitions to the next sub-goal.

\noindent\textbf{Snapshot-Conditioned Control.}
We unmask the long-horizon snapshot. We still apply the consistent per-body sampling to determine which body or object components are revealed. All short-horizon preview are fully masked. 

\noindent\textbf{Trajectory-Conditioned Control.}
We unmask the short-horizon preview. Following the same per-body sampling, we reveal only a subset of the joint or object components. The long-horizon snapshot goal is retained.

\noindent\textbf{Contact-Conditioned Control.}
Contact goals are implemented as a special case of snapshot conditioning in which we reveal only contact-related information. Specifically, we unmask the contact entries of $\boldsymbol C_t$, the associated signed-distance fields $\boldsymbol D_t$ (defined in Sec.~\ref{sec:state}), and the relevant human body parts. To avoid ambiguity in the target, we additionally unmask the object pose in the snapshot frame.

\noindent\textbf{Multi-Goal Chaining.}
For multi-goal chains, we extract data by concatenating different data sequences. Specifically, we canonicalize each subsequent first frame with respect to the previous last frame. Canonicalization is performed by aligning the human root position (excluding height), and heading, \ie, rotation around the vertical $z$–axis only, rather than the full $SO(3)$ orientation.
Because this transformation is applied with respect to the human frame only, the object frame may become partially misaligned after canonicalization. As a result, we do not expect the policy to perfectly satisfy all chained goals, especially when object-relative alignment becomes extremely inconsistent. Nevertheless, the presence of a long horizon makes the policy possibly compensate for canonicalization artifacts.

\section{Additional Details on Methodology} \label{sec:reward}

This section expands the reward and loss formulations, as well as additional details for the three stages of our framework: (\textbf{I}) InterMimic+ expert training (extending Sec.~\ref{sec:teacher}), (\textbf{II}) variational distillation (extending Sec.~\ref{sec:distill}), and (\textbf{III}) RL post-training (extending Sec.~\ref{sec:finetune}).

\subsection{InterMimic+: Full-Reference Imitation Expert}

\noindent\textbf{Reference-Free Reward for Expert.} Here we introduce the detailed formulation of the hand reward $r_\mathrm{h}$. Let \(\boldsymbol{p}_T\) denote the position of the thumb fingertip and \(\{\boldsymbol{p}_j\}_{j \in S}\) the positions of the other fingertips, with \(\boldsymbol{q}_T\) and \(\{\boldsymbol{q}_j\}_{j \in S}\) being their respective nearest surface points on the object. 
We define unit bearing vectors from the object surface toward the fingertips as \(\boldsymbol{u}_T = (\boldsymbol{p}_T {-} \boldsymbol{q}_T)/\|\boldsymbol{p}_T {-} \boldsymbol{q}_T\|\) and \(\boldsymbol{u}_j = (\boldsymbol{p}_j {-} \boldsymbol{q}_j)/\|\boldsymbol{p}_j - \boldsymbol{q}_j\|\), \(j \in S\). 
The reward is defined as \(r_{\mathrm{h}} = \exp(-w_{\mathrm{h}} e_{\mathrm{h}})\), where \(e_{\mathrm{h}} = 1 {-} \frac{1}{|S|} \sum_{j \in S} \frac{1 - \boldsymbol{u}_T^\top \boldsymbol{u}_j}{2}\), and \(w_{\mathrm{h}}\) increases as the hand-object distance decreases, activating only when the reference indicates an upcoming interaction. 
This reward encourages all five fingers to maximize upcoming surface contact with the object.

\subsection{InterPrior: Variational Distillation}

Here we introduce the formulation for our proposed losses for variational Distillation.
Let $\boldsymbol{\mu}_{p,t}$ and $\boldsymbol{\Sigma}_{p,t}$ denote the prior’s mean and covariance at time $t$,
\ie,
$\mathcal{N}(\boldsymbol{\mu}_{p,t}, \boldsymbol{\Sigma}_{p,t}) \equiv p_\psi(\boldsymbol{z}_t \mid \boldsymbol{x}_{t-\ell:t}, \mathcal{G}_t)$.

\noindent(\textbf{I}) \emph{Scale loss.}
We regularize the prior mean to lie on the unit hypersphere. This is to prevent the output mean from collapsing or exploding, with the use of latent normalization:
\[
\mathcal{L}_{\text{scale}}
= \mathbb{E}_t\bigl[\bigl(\|\boldsymbol{\mu}_{p,t}\|_2 - 1\bigr)^2\bigr].
\]

\noindent(\textbf{II}) \emph{Temporal consistency loss.}
To obtain a smooth latent prior over time, we use $\mathcal{L}_{\text{tc}}$ to penalize changes in the prior distribution across consecutive timesteps using the squared 2‑Wasserstein distance between Gaussians.

\noindent(\textbf{III}) \emph{Goal reconstruction loss.}
The decoder includes an additional head that predicts future goal features conditioned on the latent.
Let $\widehat{\boldsymbol{y}}_{t+k}$ denote the predicted goal at offset $k$ and
$\boldsymbol{m}_{t+k}$ the input mask used to construct the masked residual goal.
We train this head to complete the \emph{masked} entries of the goal, \ie, those that were hidden from the policy input.
Formally, the goal reconstruction loss is
\[
\mathcal{L}_{\text{goal}}
= \mathbb{E}_{t,k}\bigl[
\bigl\| \bigl(\mathbf{1} - \boldsymbol{m}_{t+k}\bigr)
       \odot \bigl(\widehat{\boldsymbol{y}}_{t+k} - \boldsymbol{y}_{t+k}\bigr)
\bigr\|_2^2
\bigr],
\]
where $\odot$ denotes element-wise multiplication and $\mathbf{1}$ is an all-ones vector.
This loss encourages the latent $\boldsymbol{z}_t$ to capture intent and context sufficient to reconstruct the missing parts of the goal, given only the visible subset provided by the mask. In practice, we reconstruct short future with $k=1$.

\subsection{InterPrior: Post-Training Beyond Reference}

\noindent\textbf{Get-Up Training.}
To learn the get-up behavior, in addition to the new learnable token as discussed in Sec.~\ref{sec:finetune}, we introduce an auxiliary reward that becomes active, with episodes initialized from a fallen state. The reward encourages both elevation of the pelvis and reorientation of the torso toward an upright configuration:
\begin{equation}
    r^{\text{getup}} =
        w_{\text{height}} \, \sigma\!\bigl(h_t - h_{\text{target}}\bigr) \;+\;
        w_{\text{upright}} \, \sigma\!\bigl(\mathbf{n}_t \cdot \mathbf{n}_{\text{up}}\bigr),
\end{equation}
where $h_t$ is the pelvis height, $h_{\text{target}}$ is set as $0.7$, $\mathbf{n}_t$ is the torso’s up vector, $\mathbf{n}_{\text{up}}$ is the world up direction, and $\sigma(\cdot)$ denotes a clipped linear shaping function. 

\noindent\textbf{Distributed Training.}
To mitigate catastrophic forgetting, we divide the parallel simulation environments into three groups:
(\textbf{I}) \textit{RL environments}, optimized solely with the post-training reward $r^{\text{PT}}_t$;
(\textbf{II}) \textit{Distillation environments}, optimized using the ELBO objective and supervised by the expert policy, as described in Sec.~\ref{sec:distill}.
The policy parameters are shared across all environments. Gradients are aggregated synchronously to update the shared policy.  

\noindent\textbf{Mask Prompt Engineering during Inference.}
To further enhance robustness during inference without additional learning, we apply lightweight \emph{mask-based prompting} over the goal specification $\mathcal{G}_t$ (Sec.~\ref{sec:state}):
(\textbf{I}) When following a trajectory and the state lags behind, we remove the trajectory goal but redefine the nearest waypoint as the snapshot goal. 
(\textbf{II}) For snapshot goals with distant target joints (\textgreater1\,m), we retain only the root translation goal while masking out all other components, prompting locomotion before fine manipulation. 
(\textbf{III}) When human-object targets are contradictory, \eg, both are moving but no grasp is established, we set the human root goal to the current object position while maintaining root height, masking all other joints. This encourages natural re-approach and regrasping behaviors.
These inference-time edits operate solely on the goal $\mathcal{G}_t$, while the policy parameters remain fixed.

\noindent\textbf{Finetuning on Additional HOI Datasets.}
The same finetuning mechanism naturally extends to absorbing new interaction datasets. Given any additional HOI corpus (\eg, BEHAVE~\cite{bhatnagar22behave} or HODome~\cite{zhang2023neuraldome} in Sec.~\ref{sec:exp}), states from such new dataset are treated as additional sources of long-horizon goals and initializations for RL rollouts, while the distillation group continues to regularize the policy toward the original prior. This allows InterPrior to incrementally acquire new object categories and interaction styles without retraining from scratch.

\section{Implementation Details} \label{sec:training}

This section summarizes key implementation details, including network configurations, hyperparameters, randomization settings used for expert training, and additional techniques used during G1 training for sim-to-sim experiments.

\noindent\textbf{PPO Setup.}
For both the expert and RL finetuning stages, we use PPO with generalized advantage estimation (GAE) and a clipped surrogate objective, and train with Adam. Following common practice~\cite{peng2018deepmimic}, we keep the PPO discount factor $\gamma$, GAE parameter $\lambda$, clip ratio, and entropy regularization as shown in Table~\ref{tab:ppo_hype}, and apply gradient clipping. 

\noindent\textbf{InterMimic+: Full-Reference Imitation Expert.}
The InterMimic+ expert policy and critic are MLPs with three hidden layers of sizes $(1024, 1024, 512)$, using ReLU activations. Actor and critic are parameterized separately, and the critic outputs a scalar value with full observation and reference as input. Please refer to~\cite{xu2025intermimic} for more details.

\noindent\textbf{InterPrior: Variational Distillation.}
The encoder and decoder used for variational distillation share the same MLP backbone with hidden sizes $(1024, 1024, 512)$. 
The prior $p_\psi$ is implemented as a 4-layer Transformer encoder with 4 attention heads, a latent dimension of 512, and a feedforward width of 1024. 
For the distillation objective (Sec.~\ref{sec:reward}), we use unit weight for the action reconstruction loss, and assign a weight of $10^{-3}$ to all auxiliary terms (goal reconstruction, scale loss, and temporal consistency loss). 
The KL regularizer follows a $\beta$-VAE style schedule: the KL weight $\beta$ is annealed from $10^{-3}$ to $1.0$ over the course of training. 
We first perform 500 epochs of warm-up using only teacher-controlled rollouts, and then gradually increase the fraction of student-controlled rollouts~\cite{ross2011reduction} until epoch $10,000$, at which point 95\% of environments are driven by the student policy while the remaining 5\% always use the teacher for fresh expert trajectories.

\noindent\textbf{InterPrior: Post-Training Beyond Reference.}
For the post-training stage, we retain the same loss weights used for the distillation branch, and combine with the PPO loss weights specified in Table~\ref{tab:ppo_hype} for the RL branches.

\noindent\textbf{Inference Efficiency.} The runtime breakdown is: observation 20.16,ms, physics 19.02,ms, policy inference 0.43,ms, \textit{SDF} 0.134,ms, and other overheads 0.057,ms, highlighting the policy’s potential for real-world deployment.

\begin{table}
    \caption{Hyperparamters for training teacher and student policies.}
  \begin{tabular}{l|l}
    \toprule
    Hyperparameters & value \\
    \midrule
    Discount factor $\gamma$    & 0.99\\
    Generalized advantage estimation $\lambda$    & 0.95\\
    Learning rate & 2e-5 \\
    Action loss weight & 1 \\
    Critic loss weight & 5 \\
    Action bounds loss weight & 10 \\
    Minibatch size & 16384 \\
    Horizon length $H$ & 32 \\
    Maximum episode length & 300 \\
  \bottomrule
\end{tabular}
\label{tab:ppo_hype}
\end{table}

\renewcommand{\arraystretch}{1.3}
\begin{table}
\caption{\textbf{Additional reward terms for G1} used in Stage~I expert training. 
Here, $\boldsymbol{\tau}$ denotes the vector of joint torques with elementwise limits $[\boldsymbol{\tau}_{\min}, \boldsymbol{\tau}_{\max}]$; 
$\boldsymbol{q}$ and $\dot{\boldsymbol{q}}$ are joint degrees and velocities with limits $[\boldsymbol{q}_{\min}, \boldsymbol{q}_{\max}]$; 
$\boldsymbol{a}_t$ is the control action at time $t$; 
$\boldsymbol{\omega}$ and $\boldsymbol{v}$ are the base (root) angular and linear velocities; 
$F^{\text{feet}}_z$ is the vertical ground-reaction force at the feet; 
$\boldsymbol{v}^{\text{feet}}$ is the tangential (ground-plane) velocity of the feet; 
$d_{\text{feet}}$ is the horizontal distance between the two feet, with desired bounds $[d_{\min}, d_{\max}]$; 
$\boldsymbol{g}^{\text{feet}}_{xy}$ is the projection of the gravity direction onto the foot frame’s ground plane; 
$\mathds{1}(\cdot)$ and $\mathds{1}_{\text{termination}}$ are indicator functions. 
All norms $\|\cdot\|$ and $\|\cdot\|_2$ are Euclidean.}
\label{tab:reward_stage1}
\centering
\resizebox{\linewidth}{!}{
\begin{tabular}{l c c}
\toprule[0.95pt]
\textsc{Term} & \textsc{Expression} & \textsc{Weight} \\
\midrule[0.6pt]
\multicolumn{3}{l}{\textit{\textbf{Penalty:}}} \\
\midrule[0.6pt]
Torque limits & $\mathds{1}(\boldsymbol{\tau} \notin [\boldsymbol{\tau}_{\min}, \boldsymbol{\tau}_{\max}])$ & $2$ \\
DoF position limits & $\mathds{1}(\boldsymbol{q} \notin [\boldsymbol{q}_{\min}, \boldsymbol{q}_{\max}])$ & $5$ \\
Energy & $\|\boldsymbol{\tau} \odot \dot{\boldsymbol{q}}\|$ & $10^{-4}$ \\
Termination & $\mathds{1}_{\text{termination}}$ & $-30$ \\
\midrule[0.6pt]
\multicolumn{3}{l}{\textit{\textbf{Regularization:}}} \\
\midrule[0.6pt]
DoF velocity & $\|\dot{\boldsymbol{q}}\|_2^2$ & $4\times10^{-4}$ \\
Action rate & $\|\boldsymbol{a}_t\|_2^2$ & $0.1$ \\
Torque & $\|\boldsymbol{\tau}\|$ & $2\times10^{-3}$ \\
Angular velocity & $\|\boldsymbol{\omega}\|^2$ & $0.01$ \\
Base velocity & $\|\boldsymbol{v}\|^2$ & $0.1$ \\
Foot slip & $\mathds{1}(F^{\text{feet}}_z > 5.0)\cdot \sqrt{\|\boldsymbol{v}^{\text{feet}}\|}$ & $0.03$ \\
Feet distance reward & 
\makecell{
$\frac{1}{2}\exp\left(-100\left|\max(d_{\text{feet}} - d_{\min}, -0.5)\right|\right)$\\[2pt]
$\quad +~\frac{1}{2}\exp\left(-100\left|\max(d_{\text{feet}} - d_{\max}, 0)\right|\right)$
}
& $0.5$ \\
Feet orientation & $\sqrt{\|\boldsymbol{g}^{\text{feet}}_{xy}\|}$ & $1$ \\
\bottomrule[0.95pt] 
\end{tabular}
}
\end{table}

\begin{table}
\caption{\textbf{Range of dynamics randomization.}  
``default'' refers to the parameter value from the unitree G1 official 29DoF model.  
$v_{xy}$ is the planar (horizontal) push velocity.}
\label{table:dr}
\centering
\renewcommand{\arraystretch}{1.2}
\resizebox{0.8\columnwidth}{!}{
\begin{tabular}{l c}
\toprule[0.95pt]
\textbf{Term} & \textbf{Range / Value} \\
\midrule[0.6pt]
\multicolumn{2}{l}{\textit{\textbf{Dynamics randomization}}} \\
\midrule[0.6pt]
Friction coefficient & $\mathcal{U}(1.0,\, 3.0)$ \\
Base CoM offset & $\mathcal{U}(-0.05,\, 0.05)\ \text{m}$ \\
Base mass offset & $\mathcal{U}(-3.0,\, 3.0)\ \text{kg}$ \\
P gain scaling & $\mathcal{U}(0.8,\, 1.2)\times\text{default}$ \\
D gain scaling & $\mathcal{U}(0.8,\, 1.2)\times\text{default}$ \\
\midrule[0.6pt]
\multicolumn{2}{l}{\textit{\textbf{External perturbation}}} \\
\midrule[0.6pt]
Push robot & interval $= 4$\,s,\quad $v_{xy} = 1$\,m/s \\
\bottomrule[0.95pt]
\end{tabular}
}
\end{table}

\begin{figure*}
  \centering
  \includegraphics[width=\textwidth]{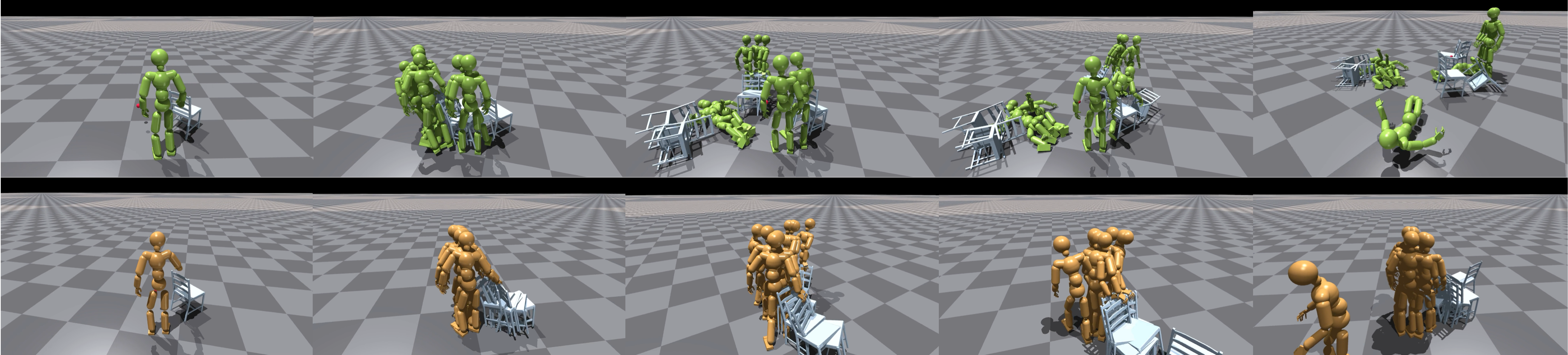}
  \caption{\textbf{Additional qualitative comparisons} with baseline method~\cite{tessler2024maskedmimic,tessler2025maskedmanipulator} (\textbf{Top}). Our InterPrior shows higher success rate under the same task goal.}
  \label{fig:failure}
\end{figure*}

\section{Additional Experimental Results} \label{sec:add_exp}

In this section, we introduce metric details, provide supplementary qualitative results, and discuss failure cases.

\noindent\textbf{Additional Details on Evaluation Metrics.}
For \emph{trajectory-following} tasks, we evaluate the policy at each timestep by comparing the rollout state with the corresponding reference, and compute pose and object errors only over the unmasked components.
For \emph{snapshot goal-following} tasks, there is no time-aligned reference trajectory. Instead, we compute the error between the rollout state and the snapshot goal at every timestep and report the \emph{minimum} of this distance over the rollout. This reflects whether the policy is capable of reaching the target configuration.

\begin{figure}
  \centering
  \includegraphics[width=\columnwidth]{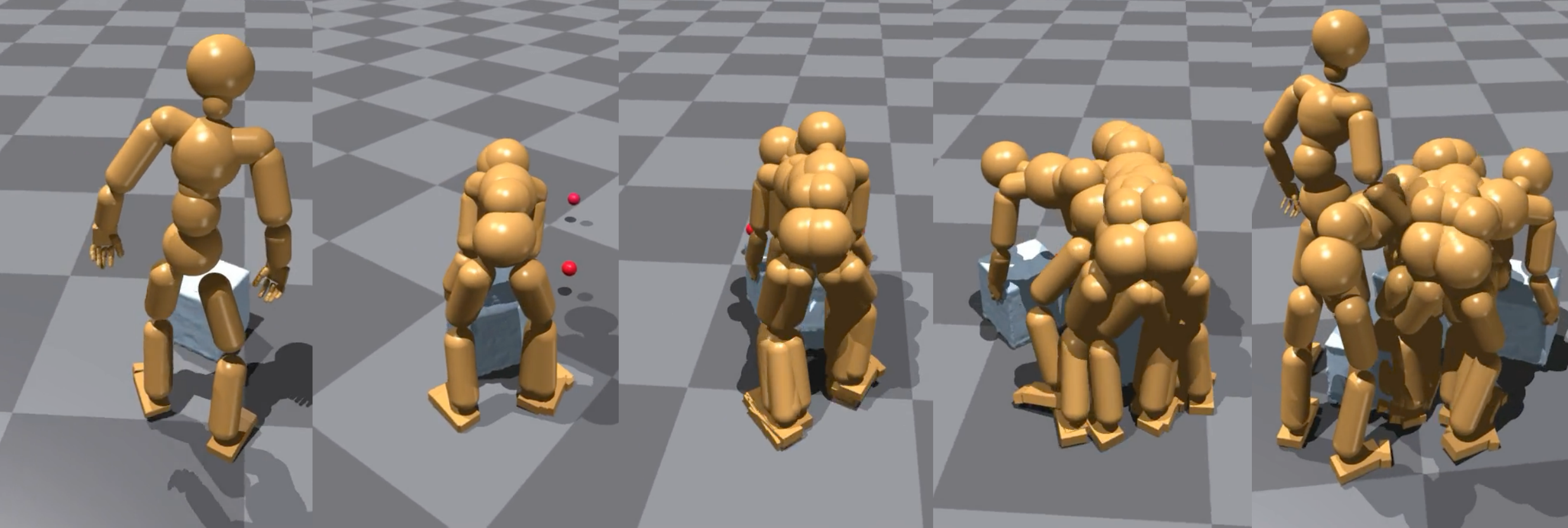}
  \caption{\textbf{Qualitative results} given the same goal. Our framework produces multiple valid yet distinct interaction trajectories.}
  \label{fig:diverse}
\end{figure}

\noindent\textbf{Diverse Behaviors Under the Same Goal.} 
Beyond the examples shown in the main paper, Figure~\ref{fig:diverse} illustrates how InterPrior behaves diversely given the same goal, showing that our learned latent space is meaningful and is able to capture diverse behaviors.

\begin{figure}
  \centering
  \includegraphics[width=\columnwidth]{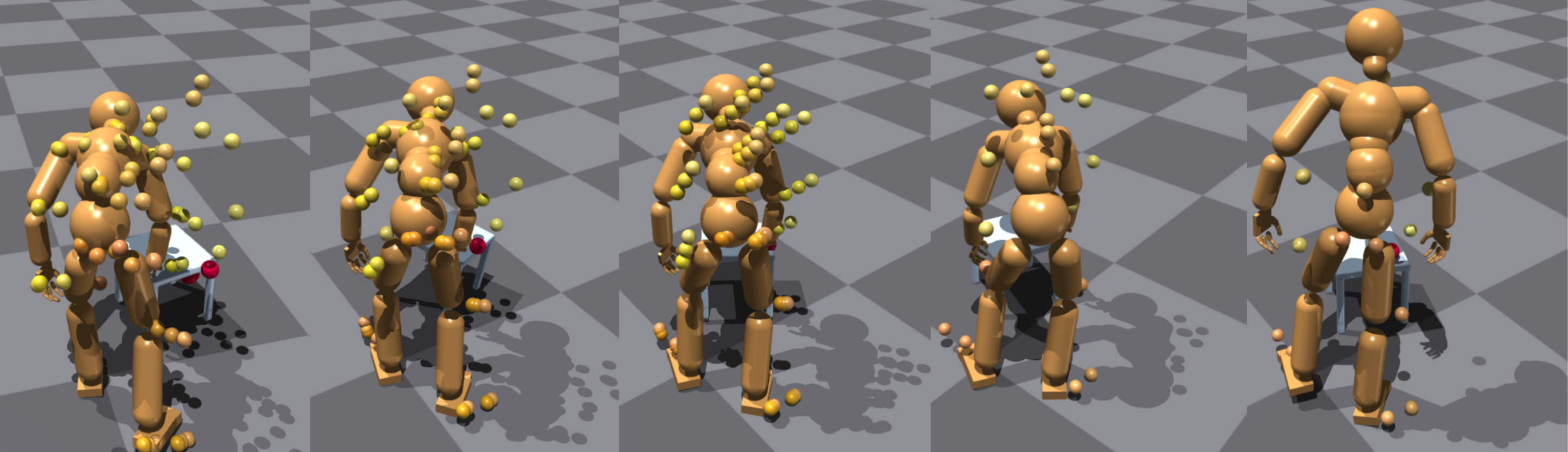}
    \caption{\textbf{Qualitative results} of InterPrior following the targets generated by InterDiff~\cite{xu2023interdiff} (\textcolor[rgb]{0.85,0.7,0.0}{\textit{yellow}} and \textcolor[rgb]{0.7,0.0,0.0}{\textit{red}} dots). InterPrior adaptively completes the task without strictly adhering to the targets, using only sparse inputs of wrist, feet, and object target.}
  \label{fig:interdiff}
\end{figure}

\noindent\textbf{Integration with Kinematic HOI Generators.}
To demonstrate that InterPrior's generalization, we integrate it with InterDiff~\cite{xu2023interdiff} that produces physically unconstrained interaction trajectories. The integration proceeds as follows: (\textbf{I}) the kinematic generator produces a 25 frames of human-object poses given the past 15 frames following~\cite{xu2023interdiff}; (\textbf{II}) we convert these sequences into our goal representation by extracting snapshot and trajectory goals; and (\textbf{III}) we feed these goals into InterPrior. The result is shown in Figure~\ref{fig:interdiff}.

\section{Discussion} \label{sec:discuss}

\noindent\textbf{Limitations and Future Work.}
InterPrior is still bounded by the coverage and quality of its training data: highly corrupted or unseen interaction patterns are not reliably recovered, and in such cases the policy often defaults to conservative strategies, maintaining balance without fully solving the task. Our model is tailored to rigid object, and we still observe occasional artifacts such as shallow interpenetrations, foot skating, or failure cases such as object drop over long rollouts. The current hand and contact representation is also not designed for fine-grained finger dexterity or in-hand manipulation. Finally, our three-stage training introduces additional complexity and hyperparameters. Future work includes expanding dataset diversity, incorporating richer hand models, and simplifying or unifying the training scheme.

\noindent\textbf{Societal and Ethical Considerations.}
InterPrior enables more general-purpose, physically grounded humanoid controller, which can be beneficial for animation, simulation, and robotics, but also raises potential risks. More capable humanoid controllers could be deployed in unsafe settings or for applications that conflict with societal norms (\eg, surveillance or coercive scenarios). We therefore encourage careful consideration of safety mechanisms, usage policies, and ethical guidelines when applying this type of model beyond controlled research environments.

\end{document}